\documentclass[11pt]{article}

\usepackage[preprint]{acl}

\usepackage{times}
\usepackage{latexsym}

\usepackage[T1]{fontenc}

\usepackage[utf8]{inputenc}

\usepackage{microtype}

\usepackage{inconsolata}

\usepackage{times}
\usepackage{soul}
\usepackage{url}
\usepackage{graphicx}
\usepackage{amsmath}
\usepackage{amsthm}
\usepackage{booktabs}
\usepackage{amsmath}
\usepackage{enumitem}
\usepackage{tabularx}
\usepackage{algorithm}
\usepackage{algorithmic}
\usepackage{caption}
\usepackage{cleveref}
\usepackage{titlesec}
\usepackage{lipsum}
\usepackage{tikz}
\usepackage{colortbl}
\usepackage{multirow}
\usepackage{array}
\usepackage{amssymb} 
\usepackage{listings}
\usepackage{tcolorbox}
\usepackage[dvipsnames]{xcolor}

\definecolor{mygray}{gray}{.9}
\definecolor{codegreen}{rgb}{0,0.6,0}
\definecolor{codegray}{rgb}{0.5,0.5,0.5}
\definecolor{codepurple}{rgb}{0.58,0,0.82}
\definecolor{backcolour}{rgb}{0.95,0.95,0.92}

\definecolor{rulecolor}{rgb}{0.1,0.1,0.5}
\definecolor{commentcolor}{rgb}{0.5,0.5,0.5}
\definecolor{featurecolor}{rgb}{0.0,0.5,0.0}

\lstdefinestyle{pddlstyle}{
    backgroundcolor=\color{backcolour},   
    commentstyle=\color{codegreen},
    keywordstyle=\color{magenta},
    numberstyle=\tiny\color{codegray},
    stringstyle=\color{codepurple},
    basicstyle=\ttfamily\footnotesize,
    breakatwhitespace=false,         
    breaklines=true,                 
    captionpos=b,                    
    keepspaces=true,                 
    numbers=left,                    
    numbersep=5pt,                  
    showspaces=false,                
    showstringspaces=false,
    showtabs=false,                  
    tabsize=2,
    language=lisp
}

\crefname{section}{\S}{\S\S}
\Crefname{section}{\S}{\S\S}

\title{What We Talk About When We Talk About LLM Planning: Evidence for Two Distinct Planning Abilities}

\author{
 \textbf{Sukai Huang},
 \textbf{Chenyuan Zhang},
 \textbf{Fucai Ke},
 \textbf{Zhixi Cai},
\\
 \textbf{Naim Rastgoo},
 \textbf{Gholamreza Haffari},
 \textbf{Hamid Rezatofighi},
\\
\\
 Faculty of Information Technology, Monash University
 \\
 \small{
   \textbf{Correspondence:} \href{mailto:sukai.huang@monash.edu}{sukai.huang@monash.edu}
 }
}

\begin{document}
\maketitle
\begin{abstract}
When LLMs exhibit uneven performance across planning tasks, these gaps are often attributed to task difficulty. We argue that this explanation is incomplete, as task-level variation may reflect distinct latent planning competencies rather than differences along a single ability spectrum. We study this question on \emph{ACPBench-Hard} by evaluating multiple LLM families under varying test-time reasoning budgets and applying a multidimensional item response theory model to uncover the latent competency structure underlying LLM planning. The analysis reveals two principal dimensions that shape planning performance: \emph{operational reasoning}, the ability to evaluate local action applicability and immediate state transitions, and \emph{structural enumeration}, the ability to reason about goal reachability and landmark structure. Operational reasoning improves under model scaling and longer reasoning traces, while structural enumeration remains comparatively insensitive. Our findings motivate competency-level evaluation of LLM planning, shifting the focus from whether models improve overall to which planning competencies improve, under what conditions, and why\footnote{code and data are provided via the Openreview platform.}.
\end{abstract}
\section{Introduction}
\label{sec:intro}
As large language models (LLMs) transition from conversational interfaces to agentic systems equipped with tool use, persistent memory, and multi-step execution capabilities, planning has become a core skill underlying agentic behavior. \citep{chowa2026language,li2026agentharness}. Symbolic planning offers a controlled testbed for this purpose: it requires finding a sequence of actions that transforms an initial state into a goal state under specified constraints, while allowing instances to be generated from formal specifications, solutions to be verified automatically, and difficulty to be varied systematically \citep{valmeekam2023planning,stein2025automating,kokel2026acpbench}.

Among recent benchmarks that use symbolic planning domains to probe LLM reasoning skills, \emph{ACPBench} \citep{kokel2025acpbench} assesses reasoning about actions, state transitions, and plan construction and validation using multiple-choice and Boolean question types. Its recent extension, \emph{ACPBench Hard} \citep{kokel2026acpbench}, moves to generative, open-ended questions and partitions planning into eight distinct subtasks. Such multi-faceted probing is common in the literature: \emph{PlanBench} \citep{valmeekam2023planning} and \emph{TRAC} \citep{he2023exploring} likewise evaluate planning through diverse question types. While these subtasks are useful for organizing evaluation and localizing failures, they should not be assumed to directly map onto the latent abilities governing model performance.

This distinction matters because recent progress in LLM reasoning has encouraged the expectation that LLM planning skill should improve with larger models, Chain-of-Thought (CoT), and stronger agent harness with tools or memory \citep{chen2025towards,xu2025toward,li2026agentharness}. Yet automated planning studies repeatedly find that LLMs remain brittle on planning benchmarks, even as models and test-time reasoning mechanisms improve \citep{kambhampati2024position,huang2025chasing}. These results are often interpreted as evidence that planning is fundamentally difficult for LLMs, or that some subtasks are simply harder than others. An open question is whether uneven task-wise performance reflects merely difficulty variation along a single planning ability, or does it reveal multiple latent competencies that respond differently to scale and reasoning traces? Without addressing this question, we cannot reconcile these two seemingly conflicting observations.

\begin{figure*}[t]
  \centering
  \includegraphics[width=0.80\textwidth]{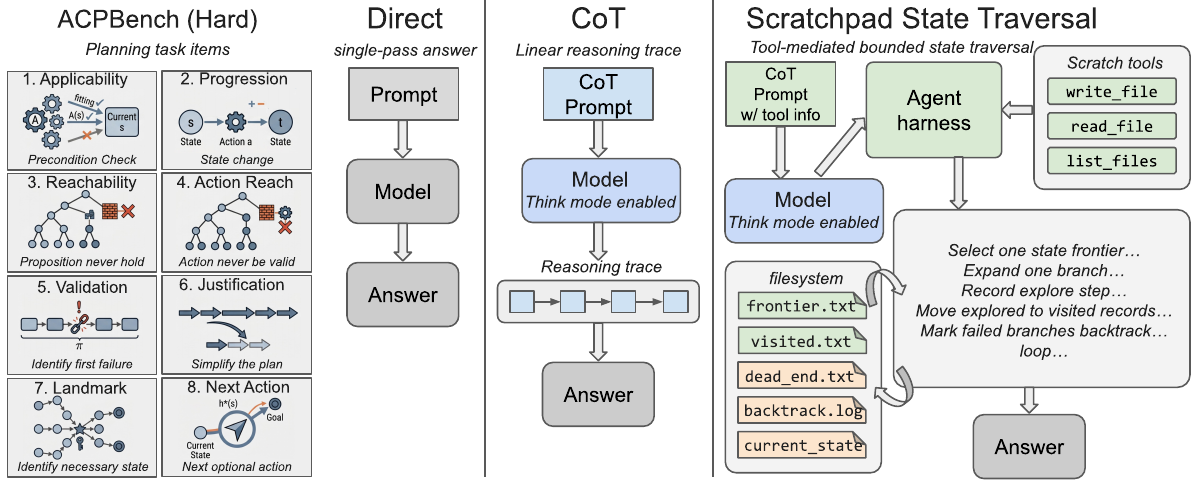}
  \caption{Evaluation design. Left: ACPBench (Hard)'s eight planning tasks (action, state, and plan levels). Right: the three inference conditions:  direct answering, CoT with linear reasoning traces, and scratchpad harness, used to probe how latent planning competencies respond to varying prompting and test-time compute scaffolding.}
  \label{fig:schematic_illustration}
\end{figure*}

To this end, we compare a family of Item Response Theory (IRT) and Multidimensional Item Response Theory (MIRT) models that vary in the number of latent dimensions and in how those dimensions combine to predict correctness. As illustrated in \Cref{fig:schematic_illustration}, our evaluation spans three LLM families, \emph{Qwen}, \emph{Gemma}, and \emph{Granite}, under three inference conditions: direct answering, Chain-of-Thought (CoT) prompting with the model's thinking mode enabled, and agent-harness scratchpad traversal. We evaluate these models on \emph{ACPBench Hard}, using model selection to determine which latent structure best explains response patterns. 

Our analysis supports the latter view. A noncompensatory two-dimensional MIRT model best explains the response data. This model reveals two interpretable dimensions: operational reasoning, reflected in applicability, progression, and validation tasks; and structural enumeration, reflected in reachability and landmark tasks. While operational reasoning improves with model size and chain-of-thought prompting, structural enumeration remains a persistent bottleneck. This asymmetric pattern holds across the model families we study.

Taken together, this work makes four contributions. First, it introduces a statistically grounded approach for interpreting LLM planning benchmark performance beyond aggregate accuracy, allowing model behavior to be analyzed through latent competency structure. Second, it provides evidence that LLM planning can decompose into distinct latent dimensions, challenging the view of planning as a single general ability. Third, it provides evidence that scaling model size and increasing test-time reasoning budget, both widely used to improve LLM reasoning, do not uniformly benefit planning, thereby identifying a systematic blind spot that aggregate benchmark scores do not expose. 


\section{Related Work and Background}

\subsection{Planning Benchmarks for LLMs}

Recent LLM planning benchmarks, such as \emph{ACPBench}~\citep{kokel2025acpbench}, \emph{ACPBench-Hard}~\citep{kokel2026acpbench}, \emph{TRAC}~\citep{he2023exploring} and \emph{ActionReasoningBench}~\citep{DBLP:conf/iclr/HandaDKSB25}, approach evaluation by decomposing end-to-end plan generation into fine-grained reasoning subtasks. 
These suites consistently report uneven task-wise performance.
ACPBench Hard attributes such variation to differences in the reasoning difficulty of individual subtasks. However, such framing implicitly treats planning as a \textbf{single underlying ability}, with subtasks occupying different points along that single dimension. This perspective overlooks a key question: whether performance gaps arise not only from \emph{how hard} a task is, but from \emph{what kind} of planning competence it requires, and whether the predefined task categories in \emph{ACPBench-Hard} correspond to distinct latent competencies. This work challenges the prevailing assumption of a single planning ability, instead identifying two competencies in LLM planning that respond asymmetrically to model scale and CoT mechanism.

\subsection{Classical Planning Structure}

Classical planning theory recognizes two modes of reasoning: transition-level reasoning (e.g., action applicability and progression) and state-space-level reasoning (e.g., reachability and landmarks)~\citep{ghallab2004automated,hoffmann2004ordered}. Modern planners do not treat these as two separate \textit{competencies}, but fuse them into unified search procedures through heuristics~\citep{DBLP:journals/ai/BonetG01}, landmarks~\citep{DBLP:journals/jair/RichterW10}, causal graphs~\citep{DBLP:journals/jair/Helmert06}, and novelty or width-based search~\citep{DBLP:conf/ecai/LipovetzkyG12,DBLP:conf/aaai/LipovetzkyG17}. Whether LLM planning behavior reflects separable, conjunctively required competencies along these lines remains an open question. Our work provides the first evidence for this two-dimensional competence structure in LLM planning behavior.

\subsection{Scaling, CoT, and Reasoning Limits}

LLM planning is a structured form of multi-step reasoning, and the prevailing assumption has long been that larger models and step-by-step reasoning via chain-of-thought (CoT) uniformly improve such capabilities~\citep{kaplan2020scaling,wei2022chain}.
Yet this view has been challenged on two fronts. \citet{DBLP:conf/icml/ChenXL0P0SLZ00T25,pu2025thoughtterminator,wu2026when} show that LLMs often experience excessive test-time compute (i.e., \emph{overthinking}) that degrades performance in reasoning tasks due to recursive self-doubt loops and error accumulation. In planning specifically, systematic evaluations tracking models from GPT-3 through GPT-4 to o1-style architectures show that even larger scales and better reasoning traces, such as those trained via Reinforcement Learning with Verifiable Rewards, cannot reliably plan~\citep{valmeekam2022large,DBLP:journals/tmlr/ValmeekamSGK25}. 
These critiques leave the performance impact of scale and reasoning traces in planning genuinely uncertain. We show that their effects are not uniform: two latent dimensions of planning competence respond asymmetrically to scale and CoT.
Recent work trains or scaffolds LLMs for planning via PDDL-aware instruction tuning \citep{verma2025teaching}, heuristic-integrated symbolic search \citep{tang2025thinksmallplansmart}, and LLM-as-formalizer \citep{DBLP:conf/aaai/HuangLC25}. We acknowledge this line of work but instead study off-the-shelf LLMs without domain-specific training in order to investigate intrinsic planning competences of LLMs.

\section{Experimental Setup}
\label{sec:experimental_setup}


ACPBench-Hard comprises eight planning-related subtasks: action applicability
(\texttt{app}), action reachability (\texttt{areach}), action
justification (\texttt{just}), landmark detection (\texttt{land}),
next-action prediction (\texttt{nexta}), state progression (\texttt{prog}),
state reachability (\texttt{reach}), and plan-goal validation
(\texttt{val}) (the left panel of \Cref{fig:schematic_illustration} gives an intuitive visual summary).
Each task contributes 130 items, yielding a total item bank of
\(J = 1{,}040\) planning instances.
We construct a binary response matrix $Y$ where $Y_{ij} = 1$ 
indicates respondent $i$ solved item $j$.

\subsection{Models and Inference Conditions}
\label{sec:models_and_conditions}

We evaluate three model families: \emph{Qwen} (Qwen3.5 0.8B, 2B, 4B, 9B, 27B, and Qwen3.8 27B), 
\emph{Gemma~4} (E2B, E4B, 12B), and \emph{Granite~3.3} (2B, 8B). 
Qwen3.5 provides the primary scale sweep while the other two test cross-family generalization. We compare three core inference conditions that span a gradient of test-time reasoning budget.

\paragraph{Direct.}
The model receives a system prompt with few-shot examples 
that show only the final answer, with no intermediate 
reasoning trace demonstrated to encourage direct generation of the final answer.

\paragraph{CoT.}
Same system prompt with few-shot 
examples that include step-by-step reasoning traces before 
the final answer~\citep{wei2022chain}, with the model's thinking mode enabled.

\paragraph{Scratchpad.}
Planning often requires branching search and backtracking from 
dead ends. Tree-of-Thought \citep{DBLP:conf/nips/YaoYZS00N23} and Graph-of-Thought \citep{DBLP:conf/aaai/BestaBKGPGGLNNH24} orchestration provide these capabilities but mainstream LLMs lack native support for them. We bridge this gap with an \emph{agent harness}: each item is solved inside the OpenCode agent, restricted to workspace file read/write tools and instructed to externalize explicit state traversal for planning tasks. Specifically, it maintains four scratch files (\texttt{frontier.md}, \texttt{visited.md}, \texttt{dead\_ends.md}, and \texttt{candidate\_answers.md}) and follows a bounded search loop to simulate state-space traversal. 

\subsection{Item Response Formulation}
\label{sec:competing_models}


Originally developed in educational testing and psychometrics, Item Response Theory (IRT) is widely used by assuming observed correctness is an indirect measurement of unobserved competence, making it a natural framework for our setting. 
We therefore cast LLM planning evaluation as an item-response problem. Each model-condition pair is a \emph{respondent}, each \emph{ACPBench-Hard} instance is an \emph{item}, and each binary correctness score, graded by the benchmark's symbolic validator against PDDL ground truth, is a \emph{response}.

\subsubsection{Unidimensional IRT}
In the simplest unidimensional IRT formulation, each respondent $i$ has a scalar latent ability $\theta_i$, and each item $j$ has a scalar difficulty $b_j$. The probability that respondent $i$ answers item $j$ correctly is modeled as
\begin{align*}
    p_{ij} &= \Pr(y_{ij}=1 \mid \theta_i, b_j) \\
           &= \sigma(\theta_i - b_j),
\end{align*}
where $\sigma(x) = (1+\exp(-x))^{-1}$. The observed response is then modeled as $y_{ij} \sim \mathrm{Bernoulli}(p_{ij})$. Under this model, correct responses are likely when a respondent's latent ability exceeds item difficulty, and unlikely otherwise. In our context, it corresponds to the hypothesis that all planning tasks are governed by a single latent planning ability, with task variation captured only by item difficulty.

\subsubsection{Multidimensional IRT}
\label{sec:pca_mirt_motivation}


Multidimensional IRT (MIRT) generalizes scalar respondent ability and item difficulty into latent vectors $\boldsymbol{\theta}i = (\theta{i1}, \ldots, \theta_{iK})$ and $\mathbf{b}j = (b{j1}, \ldots, b_{jK})$ respectively.
This formulation corresponds to the hypothesis that LLM planning may involve multiple latent competencies. Here, $\theta_{ik}$ denotes respondent $i$'s ability on latent dimension $k$, and $b_{jk}$ denotes item $j$'s difficulty on that dimension. MIRT models differ in how these latent dimensions are combined into a response probability.

A common choice is a \emph{compensatory} MIRT model, in which latent dimensions contribute additively to the logit of the response probability. For simplicity, we use a Rasch-style compensatory formulation,
\begin{equation}
    p_{ij}
    =
    \sigma\left(
        \sum_{k=1}^{K} \theta_{ik} - d_j
    \right),
\end{equation}
where $d_j$ is an item difficulty parameter. In this formulation, dimensions can compensate for one another: high ability on one dimension may offset lower ability on another because all dimensions contribute to a single additive score before the sigmoid transformation.

An alternative is a \emph{noncompensatory} MIRT model, in which each dimension contributes a separate success probability and the final response probability is their product:
\begin{equation}
    p_{ij}
    =
    \prod_{k=1}^{K}
    \sigma(\theta_{ik} - b_{jk}).
\end{equation}
Here, $b_{jk}$ denotes the difficulty of item $j$ on dimension $k$. In this formulation, a low dimension-specific success probability reduces the overall probability of a correct response, even when ability is high on another dimension. Thus, the noncompensatory model encodes the assumption that planning success may require multiple conditions to be satisfied jointly, whereas the compensatory model treats latent dimensions as partially substitutable.

\subsubsection{Model Selection}
\label{sec:irt_model_selection}

To answer our research question about the competency structure of LLM planning, we treat the choice of dimensionality ($K \in \{1,2,3,4\}$) and interaction type (compensatory vs.\ noncompensatory) as a model-selection problem. We compare candidate specifications using the Akaike Information Criterion (AIC), the Bayesian Information Criterion (BIC), and cross-validated negative log-likelihood (CV NLL). AIC and BIC are standard information criteria for comparing statistical models because they balance likelihood fit against model complexity, which is important because higher-dimensional models can improve in-sample fit simply by adding parameters. We complement these criteria with CV NLL, which directly evaluates predictive performance on held-out responses and provides a check against overfitting in more complex models. As an independent exploratory diagnostic, we also perform Principal Component Analysis (PCA) on the task-level accuracy matrix to assess the aggregate dimensionality of the response data.

Across all candidate models, we use a task-aware parameterization of item difficulty to make the learned difficulties interpretable with respect to the benchmark structure. In the unidimensional case, item difficulty is decomposed as
\[
    b_j = \alpha_{t(j)} + \epsilon_j,
\]
where $t(j)$ denotes the benchmark subtask associated with item $j$, $\alpha_{t(j)}$ is a task-level difficulty baseline shared by all items from that subtask, and $\epsilon_j$ is an item-specific residual. In the multidimensional case, the same decomposition is applied separately for each latent dimension:
\[
    b_{jk} = \alpha^{(k)}_{t(j)} + \epsilon^{(k)}_j.
\]
This formulation represents broad task-level differences explicitly while still allowing items within the same subtask to vary in difficulty.

\section{Results}
\label{sec:two_factor_model}

\subsection{Model Selection Favors a 2D Noncompensatory Structure}
\label{sec:model_selection}

\begin{table}[t]
\centering
\setlength{\tabcolsep}{3pt}
\scriptsize 
\caption{Fused held-out and information-criterion summary across IRT specifications. Train NLL is omitted for compactness; best values per column are bolded.}
\label{tab:irt_aic}
\begin{tabular}{cllcccc}
\toprule
Dims & Type & Params & CV NLL & CV Acc & AIC & BIC \\
\midrule
1 & -    & 1,089 & 0.3571 & 0.847 & 21244.7 & 21459.5 \\
2 & Comp    & 2,178 & 0.3528 & 0.847 & 23032.8 & 23462.3 \\
\textbf{2} & \textbf{Noncomp} & 2,178 & 0.3249 & 0.861 & \textbf{20978.2} & \textbf{21407.8} \\
3 & Comp    & 3,267 & 0.3518 & 0.846 & 25070.6 & 25714.9 \\
3 & Noncomp & 3,267 & 0.3208 & 0.862 & 22823.8 & 23468.1 \\
4 & Comp    & 4,356 & 0.3522 & 0.846 & 27170.9 & 28030.0 \\
4 & Noncomp & 4,356 & \textbf{0.3194} & \textbf{0.863} & 24773.3 & 25632.4 \\
\bottomrule
\end{tabular}
\end{table}

\Cref{tab:irt_aic} details parameter counts, cross-validation metrics, and information criteria across all specifications. The 2D noncompensatory model achieves the lowest overall AIC and BIC. Most of the improvement occurs when moving from the 1D to the 2D model. This transition accounts for $85.4\%$ of the total NLL reduction obtained by expanding the model from 1D to 4D, while additional dimensions provide only marginal gains and are penalized for their added complexity.
Importantly, the compensatory 2D model offers little improvement over the 1D baseline. This suggests that the two dimensions are not simply interchangeable sources of ability. Instead, weakness in either dimension can limit performance in a way that strength in the other dimension cannot fully offset.

\begin{figure}[t]
  \centering
  \includegraphics[width=0.7\columnwidth]{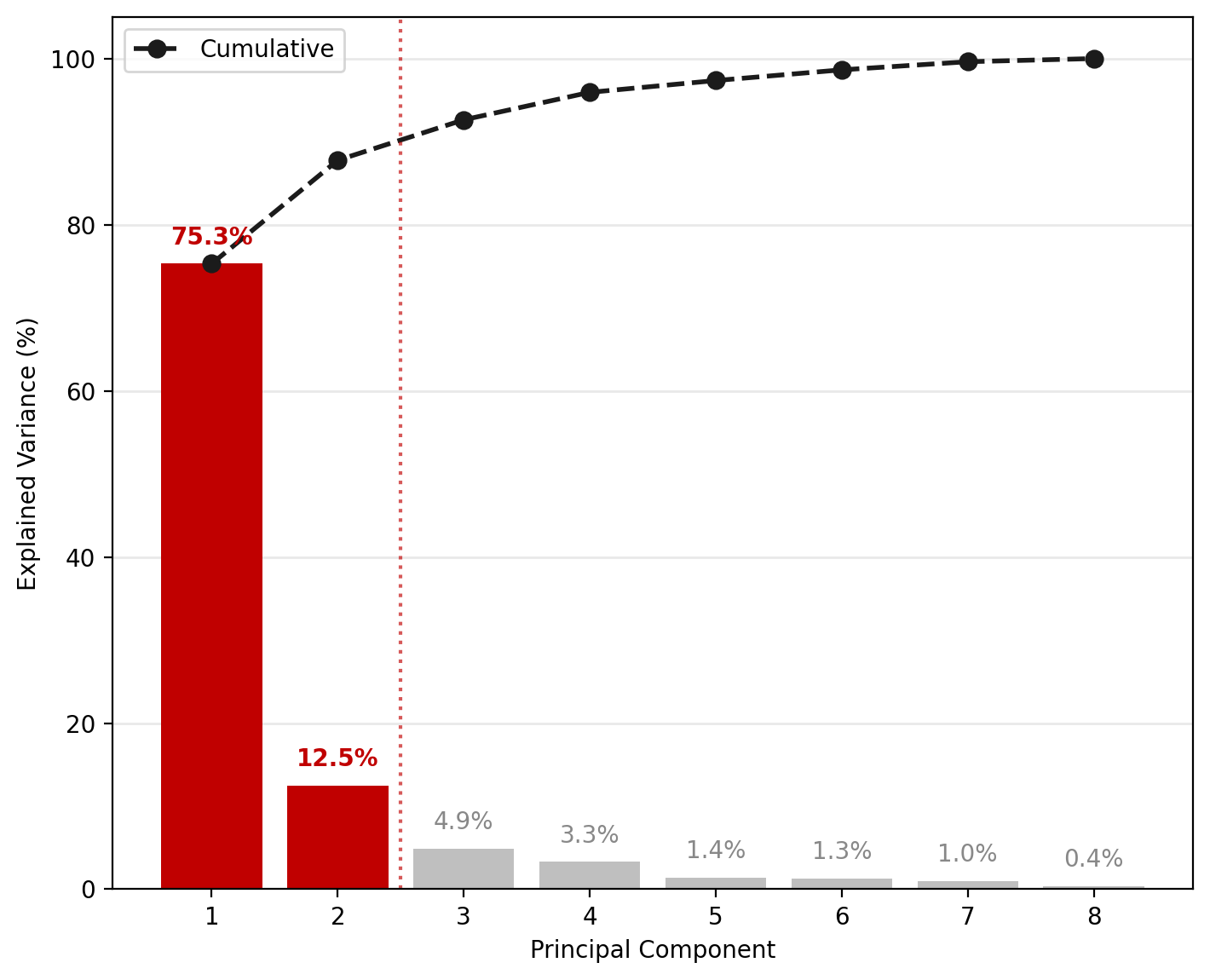}
  \caption{PCA scree plot of the task-level accuracy matrix. PC1 and PC2 
explain 88\% of variance with an elbow at component 2, 
independently corroborating the IRT dimensionality selection.}
  \label{fig:pca_scree}
\end{figure}

\Cref{fig:pca_scree} provides an auxiliary check using PCA on the task-level accuracy matrix. The first two components explain 88\% of the variance, with a clear elbow after the second component, while all remaining components each explain less than 5\%. This pattern agrees with AIC and BIC, which favor a two-dimensional noncompensatory model as the simplest model that explains the main structure of the response data. Although CV NLL and CV Acc is lowest for the 4D noncompensatory model, the improvement over the 2D model is marginal. As a result, for the remainder of the analysis, we focus on the 2D noncompensatory model because it offers the best balance between simplicity, interpretability, and predictive performance. Statistical robustness checks (permutation tests, task‑collapsed refits; \Cref{sec:appendix_robustness}) confirm that this structure is not driven by chance task labels or marginal response patterns.


For readability, we denote the task-level difficulty coordinates 
\((\alpha^{(1)}_t, \alpha^{(2)}_t)\) as \((b_1,b_2)\) when discussing task-level plots and tables. Detailed task-level difficulty coordinates are presented in \Cref{tab:task_anchors} in the appendix. \Cref{fig:task_difficulty_biplot} visualizes the tasks in the two-dimensional difficulty plane, where color denotes dimensional dominance. Red vectors correspond to tasks with stronger difficulty along Dimension~1, while blue vectors correspond to tasks with stronger difficulty along Dimension~2.

\begin{figure}[t]
  \centering
  \includegraphics[width=0.7\columnwidth]{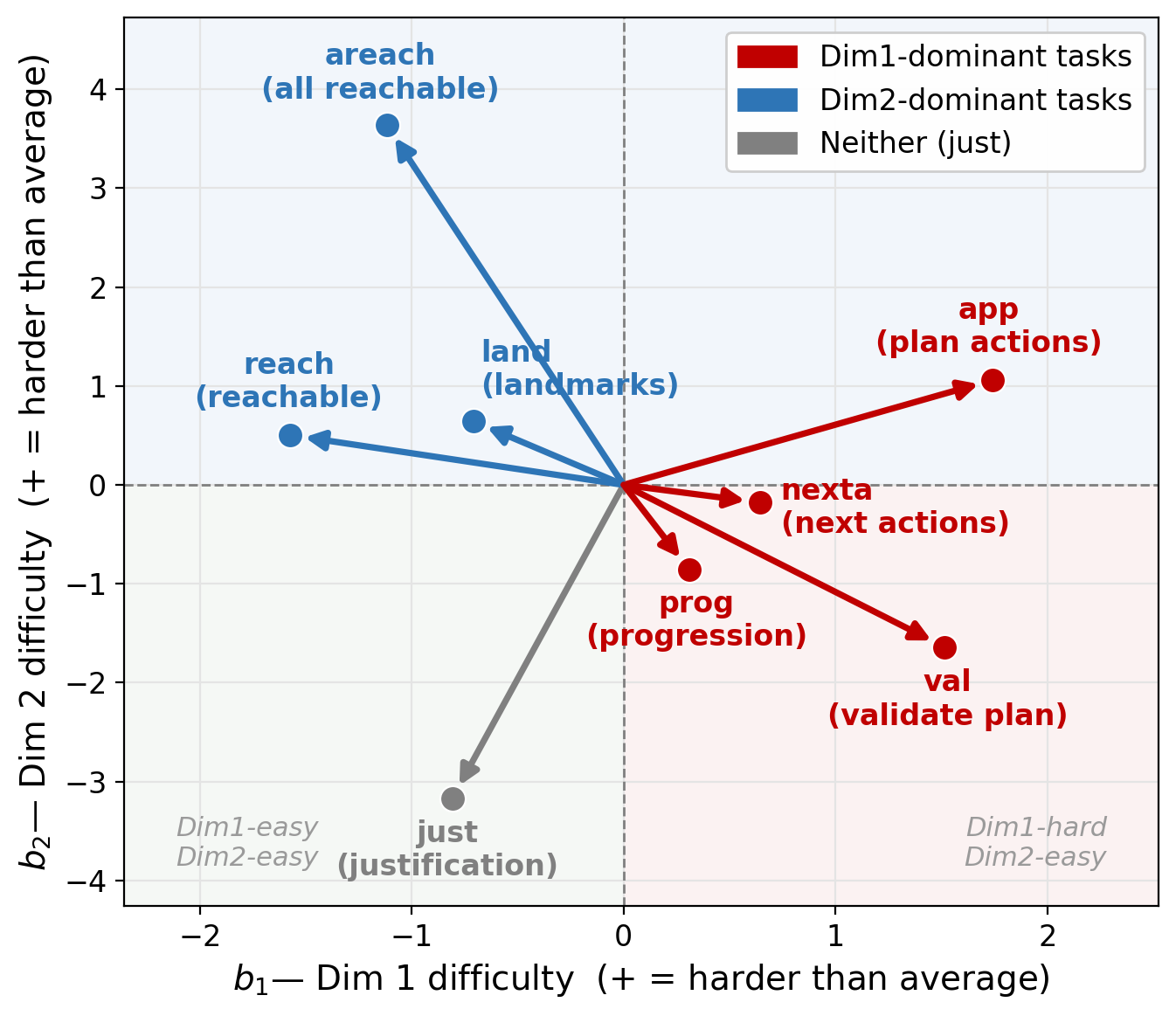}
  \caption{Biplot of task difficulty coordinates in the 2D noncompensatory model.}
  \label{fig:task_difficulty_biplot}
\end{figure}

The tasks form two clear groups in this plane. \texttt{val}, \texttt{prog}, \texttt{nexta}, and \texttt{app} have positive values on \(b_1\) that exceed their \(b_2\) values, indicating that they are primarily difficult along the first dimension; \texttt{app} is the hardest first-dimension task \((b_1 = +1.74)\) and is also above average on \(b_2\) \((+1.06)\). In contrast, \texttt{reach}, \texttt{land}, and \texttt{areach} have negative values on \(b_1\) and positive values on \(b_2\), with \texttt{areach} emerging as a clear outlier \((b_2 = +3.63)\). This pattern indicates that these tasks impose stronger difficulty along the second dimension.

The remaining tasks occupy intermediate or low-demand regions of the difficulty plane. \texttt{nexta}, which requires both local action selection and goal-directed reasoning, lies near the origin, consistent with a task that draws on both dimensions without strongly isolating either one. \texttt{just} loads weakly on both dimensions, suggesting that it places relatively low demands on either dimension for the models considered here.

We are not yet committing to what the dimensions ``mean'', instead, the following subsections characterize how the dimensions respond to model scale, inference conditions and output format.

\subsection{Scale and CoT Improve Dimension~1 but Not Dimension~2}
\label{sec:asymmetric_scaling}

If planning competence reflected a single latent ability, model scaling and test-time reasoning mechanisms should yield uniform improvements. Under the two-factor model, however, these interventions may affect the two competencies differently. These analyses test that prediction.

We evaluate three model families under three inference conditions and place
all model--condition pairs in the same \((\theta_1,\theta_2)\) space\footnote{Respondent parameters are estimated via joint calibration with shared item parameters, enabling cross comparison.}.

\begin{figure*}[t]
  \centering
  \begin{minipage}[b]{0.60\textwidth}
    \centering
    \includegraphics[width=0.95\textwidth]{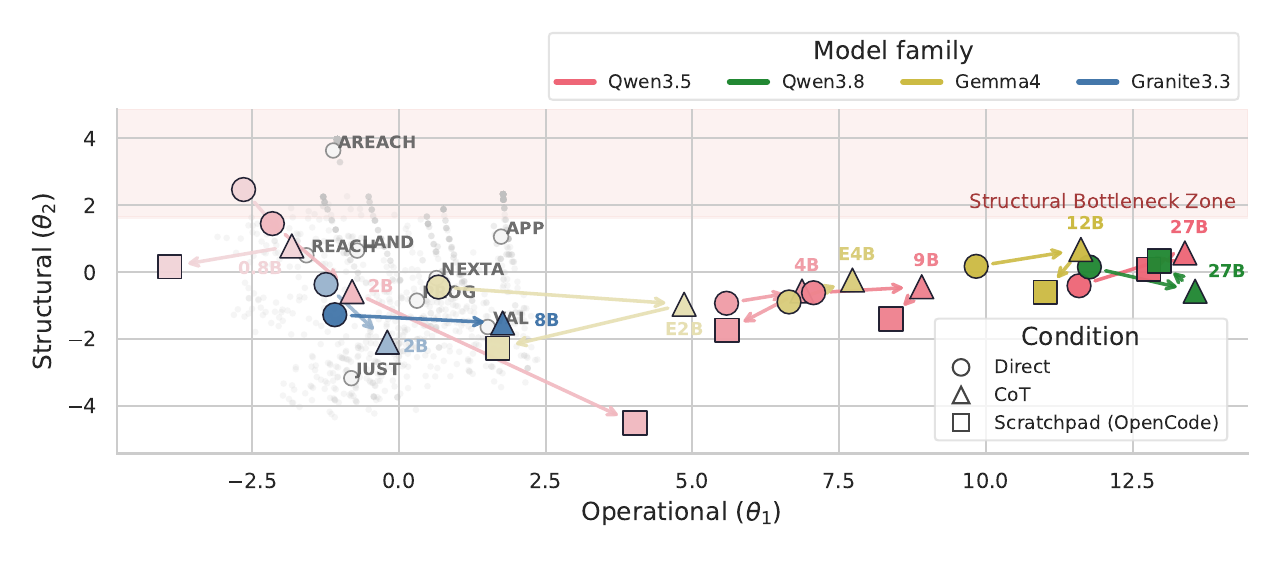}
    \caption{Asymmetric scaling of planning competence under joint calibration. 
Ability on Dimension~1 ($\theta_1$) scales with model size and prompting, while ability on Dimension~2 ($\theta_2$) flattens out.}
    \label{fig:all_family_trajectories}
  \end{minipage}
  \hfill 
  \begin{minipage}[b]{0.33\textwidth}
    \centering
    \includegraphics[width=\textwidth]{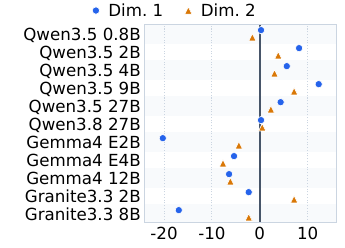}
    \caption{Targeted-minus-generic CoT accuracy (pp). Positive values favor targeted CoT.}
    \label{fig:matched_vs_mismatched_advantage}
  \end{minipage}
\end{figure*}

\Cref{fig:all_family_trajectories} plots the resulting trajectories, with arrows connecting Direct $\rightarrow$ CoT $\rightarrow$ Scratchpad for each model. The dominant movement is horizontal: as models grow larger and are given increasingly explicit reasoning scaffolds, Dimension~1 ability ($\theta_1$) increases while Dimension~2 ability ($\theta_2$) stays flat. The shaded band marks the top quartile of Dimension~2 difficulty ($b_2 \geq 1.63$), where even the strongest models remain bottlenecked; a regression analysis in \Cref{sec:appendix_scaling_regression} also supports the pattern.

The observed Dimension~2 plateau persists even with OpenCode, a widely used open-source agent harness. Scratchpad does not improve overall accuracy over CoT for seven of the eight models with a functioning agent setup. For Qwen3.5-27B, a breakdown by Dimension~2 difficulty quartile further shows that accuracy remains near floor on the highest-difficulty items (\Cref{sec:appendix_scaffold}).

\paragraph{Prompt sensitivity.}
Targeted prompting does not yield a consistent accuracy advantage across models. We compare generic CoT with a targeted CoT prompt that explicitly requests landmark, reachability, and global path reasoning. The evaluated Dimension~1-dominant subset comprises \texttt{app}, \texttt{prog}, and \texttt{val}. The Dimension~2-dominant subset comprises \texttt{land}, \texttt{reach}, and \texttt{areach}. \Cref{fig:matched_vs_mismatched_advantage} shows the accuracy difference between the prompts for each group. On Dimension~2-dominant tasks, the targeted prompt improves accuracy for most Qwen models but lowers it for all three Gemma~4 models. Its advantage also reverses from $+7.2$~pp for Granite~3.3-2B to $-2.3$~pp for Granite~3.3-8B. The direction of the effect likewise varies across model families on Dimension~1-dominant tasks.

\subsection{Externalization Failures}
\label{sec:externalization_failures}

We test whether failure signatures differ across dimensions by comparing what a linear probe can decode from hidden states with what the models themselves output. Using the model's hidden states, we train the probe to predict whether a candidate answer is valid. We compare its prediction with the model's own judgement of the same answer. Samples fall into four categories: \emph{latent absent} (probe and output both wrong), \emph{externalization failure} (probe correct, output wrong), \emph{successful} (both correct), or \emph{spurious} (probe wrong, output correct).

\begin{figure}[t]
  \centering
  \includegraphics[width=0.85\columnwidth]{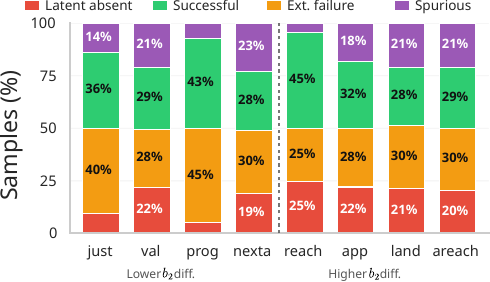}
  \caption{Probe-output failure taxonomy by task, averaged across \emph{Qwen3.5} model sizes and inference conditions, with tasks ordered by absolute Dimension~2 difficulty. }
  \label{fig:failure_taxonomy}
\end{figure}

Hidden states support more accurate validity predictions than the models' outputs (\Cref{fig:failure_taxonomy}). Mean probe accuracy is 0.66 versus 0.50 for the output judgments. Externalization failures form the largest error category at 32.0\% of all samples, compared with 18.0\% latent-absent cases. The clearest example is the Dimension~1-dominant task \texttt{prog}, which combines the highest probe accuracy (0.88) with the highest externalization rate (45.0\%).

Dimension~2-dominant tasks show both frequent externalization failures and weaker probe performance. Probe accuracy is 0.71 on \texttt{reach} and near 0.58 on \texttt{land} and \texttt{areach}. Externalization failures nevertheless remain frequent, accounting for 25.2\% of samples on \texttt{reach}, 30.2\% on \texttt{land}, and 29.7\% on \texttt{areach}. This pattern suggests that the Dimension~2 deficit extends beyond externalization alone, involving both limited linear accessibility and difficulty expressing decodable information.

This mixed signature may help explain the inconsistent gains from prompting and agent scaffolding (\Cref{fig:matched_vs_mismatched_advantage}). Prompting and scaffolding differ in form, but both work by eliciting reasoning the model is already capable of. Where the hidden states already separate valid from invalid candidates, as on \texttt{prog}, such elicitation has a correct internal judgement to draw on. On Dimension~2-dominant tasks, the hidden states misjudge far more candidates, and many output errors coincide with probe errors. This suggests that the models may largely lack the Dimension~2 capability itself, rather than holding it in a form that a better prompt could bring out. If so, elicitation alone is unlikely to close the Dimension~2 gap.

\subsection{Format Impact on Two Dimensions}
\label{sec:triplet_format_gap}

\begin{table}[t]
\centering
\scriptsize
\caption{Generative accuracy (Gen Acc), MCQ$-$Gen gap, and Bool$\to$Gen gap 
(Bool$-$Gen), sorted by 
Dim~2 difficulty $b_2$. Positive gap indicates 
inflates over Gen.}
\label{tab:format_gap}
\begin{tabular}{lrrrr}
\toprule
Task & $b_2$ & Gen Acc & MCQ$-$Gen & Bool$-$Gen \\
\midrule
\texttt{val}   & $-1.64$ & 0.455 & $+0.037$ & $+0.258$ \\
\texttt{prog}  & $-0.85$ & 0.460 & $+0.374$ & $+0.393$ \\
\texttt{reach} & $+0.50$ & 0.291 & $+0.430$ & $+0.461$ \\
\texttt{land}  & $+0.64$ & 0.245 & $+0.524$ & $+0.523$ \\
\texttt{app}   & $+1.06$ & 0.155 & $+0.633$ & $+0.662$ \\
\texttt{areach} & $+3.63$ & 0.029 & $+0.627$ & $+0.596$ \\
\midrule
\multicolumn{5}{l}{\textit{Spearman $\rho$($b_2$, 
Bool$-$Gen gap) $= 0.94$}} \\
\bottomrule
\end{tabular}
\end{table}

\Cref{sec:externalization_failures} examined the Dimension~2 deficit in a setting where candidates were already supplied. We next examine whether response format gives different impacts. Thus, we compare free-form generation with Boolean and MCQ formats to test whether supplying candidates disproportionately improves accuracy on Dimension~2-dominant tasks.

As \Cref{tab:format_gap} shows, the Bool$\to$Gen gap rises almost monotonically with Dimension~2 difficulty $b_2$ (Spearman $\rho = 0.94$). Dim2-dominant tasks suffer severe generative collapse: \texttt{areach} plunges from 0.630 (Bool, CoT) to 0.028 (Gen), and \texttt{land} and \texttt{reach} exhibit comparable drops (0.783 and 0.769 under Bool vs.\ 0.293 and 0.285 under Gen, CoT). The task with the lowest $b_2$ is the least format-sensitive: \texttt{val} shows a near-zero MCQ$-$Gen gap ($+0.04$) and the smallest Bool$-$Gen gap ($+0.26$), whereas \texttt{app}, the first-dimension task with the highest $b_2$, shows gaps as large as the Dimension~2 tasks. MCQ is equally deceptive: \texttt{areach} reaches 0.690 under MCQ (CoT) while remaining near-floor under Gen, a gap of $+0.66$. A benchmark providing candidate answers would thus severely overestimate Dimension~2 competence.

Thus we show that tasks with higher Dimension~2 difficulty benefit more from answer scaffolding. This is consistent with the structural-enumeration interpretation developed in \Cref{sec:dimension_interpretation}: free-form generation requires finding answers, whereas candidate formats supply them for evaluation.

\section{Interpreting the Two Latent Dimensions}
\label{sec:dimension_interpretation}

IRT recovers axes of statistical structure (\Cref{fig:task_difficulty_biplot}), yet the observations above converge on the same reading of the two axes. Ability on Dim~1 improves with scale and prompting, while ability on the second stays flat (\Cref{sec:asymmetric_scaling}). Second-dimension items also show more latent-absent errors (\Cref{sec:externalization_failures}). \Cref{sec:triplet_format_gap}) shows that candidate answers rescue precisely the tasks that load on Dim~2. We therefore label the first dimension \textbf{operational reasoning}, evaluating local action applicability and state transitions, and the second \textbf{structural enumeration}, reasoning about reachability and landmark structure across the state space. A pre-specified codebook analysis also agrees with this interpretation (\Cref{sec:appendix_semantic_codebook}).

\section{From Description to Manipulation}
\label{sec:interventions}

This section further validate the semantic of the two latent dimensions with targeted interventions.

\subsection{Localizing the Bottleneck: Oracle Ladder}
\label{sec:oracle_ladder}

Solving a structural item takes three stages: \emph{encode} the problem, \emph{traverse} its state space, and \emph{extract} the answer. \emph{Compile-then-enumerate} (CTE) has the model write a Python program that encodes and enumerates the state space, run in an isolated container; and \emph{oracle injection} supplies proof-verified enumeration evidence at tiers from none to complete reachable sets (\emph{None}, \emph{Partial}, \emph{Full}). Two controls share the \emph{None} prompt's length: \emph{Filler} pads the prompt with task-irrelevant text, and \emph{Distractor} supplies a perturbed enumeration that supports the wrong answer. A Reflexion-style \texttt{memory} arm is a generic-scaffold baseline \citep{DBLP:conf/nips/ShinnCGNY23}. Design details are in \Cref{sec:appendix_ladder}.

\begin{table}[t]
\centering
\setlength{\tabcolsep}{3pt}
\scriptsize
\caption{Inject enumeration info lifts the structural tasks; generic scaffolds (Memory) lift the operational tasks.}
\label{tab:ladder_results}
\begin{tabular}{lcccc}
\toprule
Task & CoT & CTE & Inject (Full) & Memory \\
\midrule
\texttt{reach}  & 56.9\% & 35.4\% & \textbf{70.0\%} & 63.1\% \\
\texttt{land}   & 53.8\% & 22.3\% & 52.3\% & \textbf{65.4\%} \\
\texttt{areach} & 6.9\%  & 7.7\%  & \textbf{10.0\%} & 7.7\% \\
\midrule
\texttt{app}    & 39.2\% & 27.7\% & 43.8\% & \textbf{46.9\%} \\
\texttt{prog}   & 76.2\% & 13.1\% & 74.6\% & \textbf{82.3\%} \\
\texttt{val}    & 93.1\% & 55.4\% & 94.6\% & \textbf{96.2\%} \\
\bottomrule
\end{tabular}
\end{table}

\paragraph{The model can read an enumeration but cannot produce one.} Under Qwen3.5-27B model, on Dim~2 heavy task \texttt{reach}, oracle injection lifts accuracy from 56.9\% (CoT) to 70.0\% (full evidence), while self-compiled enumeration (CTE, 35.4\%) falls below CoT (\Cref{tab:ladder_results}). CTE failures are mostly translation and enumeration errors in the model's own programs (68/130 on \texttt{reach}, 92/130 on \texttt{land}; \Cref{tab:cte_taxonomy}): the model encodes the problem incorrectly, so its program enumerates the wrong state space. The pipeline itself is reliable once the encoding is right. When the program's own enumeration is correct (325 of 1{,}040 items), extraction never loses it and CTE answers all of them correctly, while CoT solves only 70.3\% of the Dim~2 heavy tasks (i.e., \texttt{land/reach/areach}). The Reflexion-style \texttt{memory} arm lifts the operational tasks but leaves \texttt{areach} at floor, so generic scaffolding does not elicit structural enumeration reasoning capability of the model. Gemma-4-12B shows the same jump under injection (see \Cref{sec:appendix_ladder_results}).

\begin{table}[t]
\centering
\setlength{\tabcolsep}{3pt}
\scriptsize
\caption{Dose response on \texttt{reach}: accuracy by injected evidence tier. Controls separate evidence content from prompt volume. True evidence helps only at \emph{Full}, distractors drive accuracy below \emph{None}, and length-matched filler is indistinguishable from \emph{None}.}
\label{tab:dose_response}
\begin{tabular}{lccccc}
\toprule
Model & Filler & None & Partial & Full & Distractor \\
\midrule
Qwen3.5-27B  & 50.8\% & 55.4\% & 36.9\% & \textbf{70.0\%} & 35.8\% \\
Gemma-4-12B  & 49.2\% & 51.5\% & 36.2\% & \textbf{66.2\%} & 26.0\% \\
\bottomrule
\end{tabular}%
\end{table}
\paragraph{Evidence content, not volume, drives the effect.} \Cref{tab:dose_response} shows the dose response on \texttt{reach}. Partial evidence misleads the model below the no-evidence baseline, full evidence lifts accuracy by 13--15~pp on both models. The controls pin the mechanism on content. Distractor evidence pushes Qwen3.5-27B 22.8~pp below the no-evidence baseline (95\% CI $[-33.3, -12.2]$; Gemma $-28.5$~pp, CI $[-39.0, -17.9]$), while filler text matches the baseline.

\subsection{Dimensional Selectivity of the Effect}
\label{sec:manipulation}

Entering each arm into the joint MIRT as a new respondent (200 item-bootstrap replicates) tests the prediction directly. Oracle injection raises $\theta_2$ and leaves $\theta_1$ unmoved (full evidence: $\Delta\theta_2$ CI $[+0.22, +0.80]$, $\Delta\theta_1$ CI spans $0$). Forcing the model to produce the enumeration itself moves $\theta_2$ in the opposite direction (\texttt{cte\_matched}: $\Delta\theta_2$ CI $[-0.89, -0.38]$). Gemma-4-12B shows the same selectivity (full evidence: $\Delta\theta_2$ $[+0.47, +0.96]$; \texttt{cte\_matched}: $[-1.11, -0.69]$; \Cref{sec:appendix_ladder_results}). It thus demonstrates that Dimension~2 is moved by where enumeration happens, and only by that.

\section{Conclusion}
\label{sec:conclusion}

We examined whether LLM PDDL planning benchmark reflects a single general ability or a structured set of underlying competencies. Using IRT-based model comparison across three model families and inference conditions, we find that a 2D noncompensatory model provides the clearest account of the response patterns. The recovered dimensions corresponds to \emph{operational reasoning} and \emph{structural enumeration}. This distinction changes how we interpret progress in LLM planning. Because structural enumeration remains flat across model sizes and inference budgets, closing this gap likely requires interventions beyond scaling or longer reasoning traces. The broader lesson is that LLM planning should be evaluated not only by how much models improve, but by the structure of that improvement. Moving from aggregate accuracy to diagnostic competence profiles makes this structure visible, and our IRT framing provides a concrete tool for doing so. This perspective can guide future work from asking whether models improve overall toward identifying which planning competencies improve, under what conditions, and why.

\section{Limitations}
\label{sec:limitations}

Our findings are subject to two scope limitations that also point to
future work.

\paragraph{Model coverage.}
We jointly calibrate Qwen3.5, Qwen3.8, Gemma~4, and Granite~3.3, all of which are open-weight models with publicly available checkpoints. We do not evaluate the latest proprietary systems, such as GPT-5.5 or Opus~4.7, due to our commitment to open-source reproducibility and budget constraints.
This choice bounds the upper range of model scale we can observe, and it is possible that qualitatively different structural reasoning capabilities emerge at larger scales or under different training regimes.
However, the consistency of the two-factor structure and the asymmetric scaling pattern across three independent model families suggests that our main findings are unlikely to be artifacts of a particular model series.
Extending the analysis to larger proprietary models, where feasible, remains a natural next step.

\paragraph{Inference scaffolds and intervention scope.}
Recent progress in agentic frameworks shows that sophisticated inference harnesses integrating tool use, code execution, persistent memory, and multi-step orchestration can substantially improve LLM performance on complex reasoning tasks \citep{li2026agentharness}. Our scratchpad condition runs each item inside OpenCode, an open-source agent harness, with file-based scratchpad tools. On Qwen 3.5 27B, the harness improved overall accuracy over direct answering but not over CoT, and the bottleneck persisted. We found that in most cases, LLMs produce genuine wrong or incomplete answers, such as partial applicable-action lists and wrong reachability atoms after multi-round scratchpad bookkeeping. It implies heavy harness use may burden medium-scale LLMs with additional reasoning overhead.

\bibliography{custom}

@misc{li2026agentharness,
  title={Agent Harness Engineering: A Survey},
  author={Li, Junjie and Xiao, Xi and Zhang, Yunbei and Liu, Chen and
          Zhao, Lin and Liao, Xiaoying and Ji, Yingrui and Wang, Janet and
          Gu, Jianyang and Ge, Yingqiang and Xu, Weijie and Fang, Xi and
          Xu, Xiang and Zhao, Tianchen and Kim, Youngeun and
          Wang, Tianyang and Hamm, Jihun and Krishnaswamy, Smita and
          Huan, Jun and Reddy, Chandan},
  url={https://openreview.net/pdf?id=eONq7FdiHa},
  year={2026}
}

@article{chowa2026language,
  title={From language to action: a review of large language models as autonomous agents and tool users},
  author={Chowa, Sadia Sultana and Alvi, Riasad and Rahman, Subhey Sadi and Rahman, Md Abdur and Raiaan, Mohaimenul Azam Khan and Islam, Md Rafiqul and Hussain, Mukhtar and Azam, Sami},
  journal={Artificial Intelligence Review},
  year={2026},
  publisher={Springer}
}

@inproceedings{stein2025automating,
  title={Automating the generation of prompts for llm-based action choice in pddl planning},
  author={Stein, Katharina and Fi{\v{s}}er, Daniel and Hoffmann, J{\"o}rg and Koller, Alexander},
  booktitle={Proceedings of the International Conference on Automated Planning and Scheduling},
  volume={35},
  number={1},
  pages={250--259},
  year={2025}
}

@inproceedings{kokel2025acpbench,
  title={Acpbench: Reasoning about action, change, and planning},
  author={Kokel, Harsha and Katz, Michael and Srinivas, Kavitha and Sohrabi, Shirin},
  booktitle={Proceedings of the AAAI Conference on Artificial Intelligence},
  volume={39},
  number={25},
  pages={26559--26568},
  year={2025}
}

@inproceedings{huang2025chasing,
  title={Chasing progress, not perfection: Revisiting strategies for end-to-end llm plan generation},
  author={Huang, Sukai and Cohn, Trevor and Lipovetzky, Nir},
  booktitle={Proceedings of the International Conference on Automated Planning and Scheduling},
  volume={35},
  number={1},
  pages={204--212},
  year={2025}
}

@article{chen2025towards,
  title={Towards reasoning era: A survey of long chain-of-thought for reasoning large language models},
  author={Chen, Qiguang and Qin, Libo and Liu, Jinhao and Peng, Dengyun and Guan, Jiannan and Wang, Peng and Hu, Mengkang and Zhou, Yuhang and Gao, Te and Che, Wanxiang},
  journal={arXiv preprint arXiv:2503.09567},
  year={2025}
}

@article{xu2025toward,
  title={Toward large reasoning models: A survey of reinforced reasoning with large language models},
  author={Xu, Fengli and Hao, Qianyue and Shao, Chenyang and Zong, Zefang and Li, Yu and Wang, Jingwei and Zhang, Yunke and Wang, Jingyi and Lan, Xiaochong and Gong, Jiahui and others},
  journal={Patterns},
  volume={6},
  number={10},
  year={2025},
  publisher={Elsevier}
}

@inproceedings{he2023exploring,
  title={Exploring the capacity of pretrained language models for reasoning about actions and change},
  author={He, Weinan and Huang, Canming and Xiao, Zhanhao and Liu, Yongmei},
  booktitle={Proceedings of the 61st Annual Meeting of the Association for Computational Linguistics (Volume 1: Long Papers)},
  pages={4629--4643},
  year={2023}
}

@inproceedings{kambhampati2024position,
  title={Position: LLMs can’t plan, but can help planning in LLM-modulo frameworks},
  author={Kambhampati, Subbarao and Valmeekam, Karthik and Guan, Lin and Verma, Mudit and Stechly, Kaya and Bhambri, Siddhant and Saldyt, Lucas Paul and Murthy, Anil B},
  booktitle={Forty-first International Conference on Machine Learning},
  year={2024}
}

@article{valmeekam2023planning,
  title={On the planning abilities of large language models-a critical investigation},
  author={Valmeekam, Karthik and Marquez, Matthew and Sreedharan, Sarath and Kambhampati, Subbarao},
  journal={Advances in neural information processing systems},
  volume={36},
  pages={75993--76005},
  year={2023}
}

@inproceedings{
kokel2026acpbench,
title={{ACPB}ench Hard: Unrestrained Reasoning about Action, Change, and Planning},
author={Harsha Kokel and Michael Katz and Kavitha Srinivas and Shirin Sohrabi},
booktitle={The Fourteenth International Conference on Learning Representations},
year={2026},
url={https://openreview.net/forum?id=WIXohR7mEo}
}

@inproceedings{DBLP:conf/iclr/HandaDKSB25,
  author       = {Divij Handa and
                  Pavel Dolin and
                  Shrinidhi Kumbhar and
                  Tran Cao Son and
                  Chitta Baral},
  title        = {ActionReasoningBench: Reasoning about Actions with and without Ramification
                  Constraints},
  booktitle    = {The Thirteenth International Conference on Learning Representations,
                  {ICLR} 2025, Singapore, April 24-28, 2025},
  publisher    = {OpenReview.net},
  year         = {2025},
  url          = {https://openreview.net/forum?id=NUD03NBDOE},
  bibsource    = {dblp computer science bibliography, https://dblp.org}
}

@inproceedings{DBLP:conf/icml/ChenXL0P0SLZ00T25,
  author       = {Xingyu Chen and
                  Jiahao Xu and
                  Tian Liang and
                  Zhiwei He and
                  Jianhui Pang and
                  Dian Yu and
                  Linfeng Song and
                  Qiuzhi Liu and
                  Mengfei Zhou and
                  Zhuosheng Zhang and
                  Rui Wang and
                  Zhaopeng Tu and
                  Haitao Mi and
                  Dong Yu},
  editor       = {Aarti Singh and
                  Maryam Fazel and
                  Daniel Hsu and
                  Simon Lacoste{-}Julien and
                  Felix Berkenkamp and
                  Tegan Maharaj and
                  Kiri Wagstaff and
                  Jerry Zhu},
  title        = {Do {NOT} Think That Much for 2+3=? On the Overthinking of Long Reasoning
                  Models},
  booktitle    = {Forty-second International Conference on Machine Learning, {ICML}
                  2025, Vancouver, BC, Canada, July 13-19, 2025},
  series       = {Proceedings of Machine Learning Research},
  publisher    = {{PMLR} / OpenReview.net},
  year         = {2025},
  url          = {https://proceedings.mlr.press/v267/chen25bx.html},
  bibsource    = {dblp computer science bibliography, https://dblp.org}
}

@inproceedings{
pu2025thoughtterminator,
title={ThoughtTerminator: Benchmarking, Calibrating, and Mitigating Overthinking in Reasoning Models},
author={Xiao Pu and Michael Saxon and Wenyue Hua and William Yang Wang},
booktitle={Second Conference on Language Modeling},
year={2025},
url={https://openreview.net/forum?id=oHR862dpMC}
}

@inproceedings{
wu2026when,
title={When More is Less: Understanding Chain-of-Thought Length in {LLM}s},
author={Yuyang Wu and Yifei Wang and Ziyu Ye and Tianqi Du and Stefanie Jegelka and Yisen Wang},
booktitle={The Fourteenth International Conference on Learning Representations},
year={2026},
url={https://openreview.net/forum?id=6QDFsYxtI1}
}

@book{ghallab2004automated,
  title={Automated Planning: theory and practice},
  author={Ghallab, Malik and Nau, Dana and Traverso, Paolo},
  year={2004},
  publisher={Elsevier}
}

@article{hoffmann2004ordered,
  title={Ordered landmarks in planning},
  author={Hoffmann, J{\"o}rg and Porteous, Julie and Sebastia, Laura},
  journal={Journal of Artificial Intelligence Research},
  volume={22},
  pages={215--278},
  year={2004}
}

@inproceedings{DBLP:conf/aaai/LipovetzkyG17,
  author       = {Nir Lipovetzky and
                  Hector Geffner},
  editor       = {Satinder Singh and
                  Shaul Markovitch},
  title        = {Best-First Width Search: Exploration and Exploitation in Classical
                  Planning},
  booktitle    = {Proceedings of the Thirty-First {AAAI} Conference on Artificial Intelligence,
                  February 4-9, 2017, San Francisco, California, {USA}},
  pages        = {3590--3596},
  publisher    = {{AAAI} Press},
  year         = {2017},
  url          = {https://doi.org/10.1609/aaai.v31i1.11027},
  doi          = {10.1609/AAAI.V31I1.11027},
  bibsource    = {dblp computer science bibliography, https://dblp.org}
}

@inproceedings{DBLP:conf/ecai/LipovetzkyG12,
  author       = {Nir Lipovetzky and
                  Hector Geffner},
  editor       = {Luc De Raedt and
                  Christian Bessiere and
                  Didier Dubois and
                  Patrick Doherty and
                  Paolo Frasconi and
                  Fredrik Heintz and
                  Peter J. F. Lucas},
  title        = {Width and Serialization of Classical Planning Problems},
  booktitle    = {{ECAI} 2012 - 20th European Conference on Artificial Intelligence.
                  Including Prestigious Applications of Artificial Intelligence {(PAIS-2012)}
                  System Demonstrations Track, Montpellier, France, August 27-31 , 2012},
  series       = {Frontiers in Artificial Intelligence and Applications},
  pages        = {540--545},
  publisher    = {{IOS} Press},
  year         = {2012},
  url          = {https://doi.org/10.3233/978-1-61499-098-7-540},
  doi          = {10.3233/978-1-61499-098-7-540},
  bibsource    = {dblp computer science bibliography, https://dblp.org}
}

@article{DBLP:journals/jair/Helmert06,
  author       = {Malte Helmert},
  title        = {The Fast Downward Planning System},
  journal      = {J. Artif. Intell. Res.},
  volume       = {26},
  pages        = {191--246},
  year         = {2006},
  url          = {https://doi.org/10.1613/jair.1705},
  doi          = {10.1613/JAIR.1705},
  bibsource    = {dblp computer science bibliography, https://dblp.org}
}

@article{DBLP:journals/jair/RichterW10,
  author       = {Silvia Richter and
                  Matthias Westphal},
  title        = {The {LAMA} Planner: Guiding Cost-Based Anytime Planning with Landmarks},
  journal      = {J. Artif. Intell. Res.},
  volume       = {39},
  pages        = {127--177},
  year         = {2010},
  url          = {https://doi.org/10.1613/jair.2972},
  doi          = {10.1613/JAIR.2972},
  bibsource    = {dblp computer science bibliography, https://dblp.org}
}

@article{DBLP:journals/ai/BonetG01,
  author       = {Blai Bonet and
                  Hector Geffner},
  title        = {Planning as heuristic search},
  journal      = {Artif. Intell.},
  volume       = {129},
  number       = {1-2},
  pages        = {5--33},
  year         = {2001},
  url          = {https://doi.org/10.1016/S0004-3702(01)00108-4},
  doi          = {10.1016/S0004-3702(01)00108-4},
  bibsource    = {dblp computer science bibliography, https://dblp.org}
}

@article{wei2022chain,
  title={Chain-of-thought prompting elicits reasoning in large language models},
  author={Wei, Jason and Wang, Xuezhi and Schuurmans, Dale and Bosma, Maarten and Xia, Fei and Chi, Ed and Le, Quoc V and Zhou, Denny and others},
  journal={Advances in neural information processing systems},
  volume={35},
  pages={24824--24837},
  year={2022}
}

@article{kaplan2020scaling,
  title={Scaling laws for neural language models},
  author={Kaplan, Jared and McCandlish, Sam and Henighan, Tom and Brown, Tom B and Chess, Benjamin and Child, Rewon and Gray, Scott and Radford, Alec and Wu, Jeffrey and Amodei, Dario},
  journal={arXiv preprint arXiv:2001.08361},
  year={2020}
}

@inproceedings{valmeekam2022large,
  title={Large language models still can't plan (a benchmark for LLMs on planning and reasoning about change)},
  author={Valmeekam, Karthik and Olmo, Alberto and Sreedharan, Sarath and Kambhampati, Subbarao},
  booktitle={NeurIPS 2022 Foundation Models for Decision Making Workshop},
  year={2022}
}

@article{DBLP:journals/tmlr/ValmeekamSGK25,
  author       = {Karthik Valmeekam and
                  Kaya Stechly and
                  Atharva Gundawar and
                  Subbarao Kambhampati},
  title        = {A Systematic Evaluation of the Planning and Scheduling Abilities of
                  the Reasoning Model o1},
  journal      = {Trans. Mach. Learn. Res.},
  volume       = {2025},
  year         = {2025},
  url          = {https://openreview.net/forum?id=FkKBxp0FhR},
  bibsource    = {dblp computer science bibliography, https://dblp.org}
}

@inproceedings{DBLP:conf/nips/ShinnCGNY23,
  author       = {Noah Shinn and
                  Federico Cassano and
                  Ashwin Gopinath and
                  Karthik Narasimhan and
                  Shunyu Yao},
  editor       = {Alice Oh and
                  Tristan Naumann and
                  Amir Globerson and
                  Kate Saenko and
                  Moritz Hardt and
                  Sergey Levine},
  title        = {Reflexion: language agents with verbal reinforcement learning},
  booktitle    = {Advances in Neural Information Processing Systems 36: Annual Conference
                  on Neural Information Processing Systems 2023, NeurIPS 2023, New Orleans,
                  LA, USA, December 10 - 16, 2023},
  year         = {2023},
  url          = {http://papers.nips.cc/paper\_files/paper/2023/hash/1b44b878bb782e6954cd888628510e90-Abstract-Conference.html},
  bibsource    = {dblp computer science bibliography, https://dblp.org}
}

@inproceedings{DBLP:conf/nips/YaoYZS00N23,
  author       = {Shunyu Yao and
                  Dian Yu and
                  Jeffrey Zhao and
                  Izhak Shafran and
                  Tom Griffiths and
                  Yuan Cao and
                  Karthik Narasimhan},
  editor       = {Alice Oh and
                  Tristan Naumann and
                  Amir Globerson and
                  Kate Saenko and
                  Moritz Hardt and
                  Sergey Levine},
  title        = {Tree of Thoughts: Deliberate Problem Solving with Large Language Models},
  booktitle    = {Advances in Neural Information Processing Systems 36: Annual Conference
                  on Neural Information Processing Systems 2023, NeurIPS 2023, New Orleans,
                  LA, USA, December 10 - 16, 2023},
  year         = {2023},
  url          = {http://papers.nips.cc/paper\_files/paper/2023/hash/271db9922b8d1f4dd7aaef84ed5ac703-Abstract-Conference.html},
  bibsource    = {dblp computer science bibliography, https://dblp.org}
}

@inproceedings{DBLP:conf/aaai/BestaBKGPGGLNNH24,
  author       = {Maciej Besta and
                  Nils Blach and
                  Ales Kubicek and
                  Robert Gerstenberger and
                  Michal Podstawski and
                  Lukas Gianinazzi and
                  Joanna Gajda and
                  Tomasz Lehmann and
                  Hubert Niewiadomski and
                  Piotr Nyczyk and
                  Torsten Hoefler},
  editor       = {Michael J. Wooldridge and
                  Jennifer G. Dy and
                  Sriraam Natarajan},
  title        = {Graph of Thoughts: Solving Elaborate Problems with Large Language
                  Models},
  booktitle    = {Thirty-Eighth {AAAI} Conference on Artificial Intelligence, {AAAI}
                  2024, Thirty-Sixth Conference on Innovative Applications of Artificial
                  Intelligence, {IAAI} 2024, Fourteenth Symposium on Educational Advances
                  in Artificial Intelligence, {EAAI} 2014, February 20-27, 2024, Vancouver,
                  Canada},
  pages        = {17682--17690},
  publisher    = {{AAAI} Press},
  year         = {2024},
  url          = {https://doi.org/10.1609/aaai.v38i16.29720},
  doi          = {10.1609/AAAI.V38I16.29720},
  bibsource    = {dblp computer science bibliography, https://dblp.org}
}

@inproceedings{kwon2023efficient,
  title={Efficient Memory Management for Large Language Model Serving with PagedAttention},
  author={Woosuk Kwon and Zhuohan Li and Siyuan Zhuang and Ying Sheng and Lianmin Zheng and Cody Hao Yu and Joseph E. Gonzalez and Hao Zhang and Ion Stoica},
  booktitle={Proceedings of the ACM SIGOPS 29th Symposium on Operating Systems Principles},
  year={2023}
}

@article{verma2025teaching,
  title={Teaching llms to plan: Logical chain-of-thought instruction tuning for symbolic planning},
  author={Verma, Pulkit and La, Ngoc and Favier, Anthony and Mishra, Swaroop and Shah, Julie A},
  journal={arXiv preprint arXiv:2509.13351},
  year={2025}
}

@misc{tang2025thinksmallplansmart,
      title={Think Small, Plan Smart: Minimalist Symbolic Abstraction and Heuristic Subspace Search for LLM-Guided Task Planning}, 
      author={Junfeng Tang and Yuping Yan and Zihan Ye and Zhenshou and Song and Zeqi Zheng and Yaochu Jin},
      year={2025},
      eprint={2501.15214},
      archivePrefix={arXiv},
      primaryClass={cs.RO},
      url={https://arxiv.org/abs/2501.15214}, 
}

@inproceedings{DBLP:conf/aaai/HuangLC25,
  author       = {Sukai Huang and
                  Nir Lipovetzky and
                  Trevor Cohn},
  editor       = {Toby Walsh and
                  Julie Shah and
                  Zico Kolter},
  title        = {Planning in the Dark: LLM-Symbolic Planning Pipeline Without Experts},
  booktitle    = {Thirty-Ninth {AAAI} Conference on Artificial Intelligence, Thirty-Seventh
                  Conference on Innovative Applications of Artificial Intelligence,
                  Fifteenth Symposium on Educational Advances in Artificial Intelligence,
                  {AAAI} 2025, Philadelphia, PA, USA, February 25 - March 4, 2025},
  pages        = {26542--26550},
  publisher    = {{AAAI} Press},
  year         = {2025},
  url          = {https://doi.org/10.1609/aaai.v39i25.34855},
  doi          = {10.1609/AAAI.V39I25.34855},
  bibsource    = {dblp computer science bibliography, https://dblp.org}
}

@inproceedings{DBLP:conf/nips/ValmeekamMHSK23,
  author       = {Karthik Valmeekam and
                  Matthew Marquez and
                  Alberto Olmo Hernandez and
                  Sarath Sreedharan and
                  Subbarao Kambhampati},
  editor       = {Alice Oh and
                  Tristan Naumann and
                  Amir Globerson and
                  Kate Saenko and
                  Moritz Hardt and
                  Sergey Levine},
  title        = {PlanBench: An Extensible Benchmark for Evaluating Large Language Models
                  on Planning and Reasoning about Change},
  booktitle    = {Advances in Neural Information Processing Systems 36: Annual Conference
                  on Neural Information Processing Systems 2023, NeurIPS 2023, New Orleans,
                  LA, USA, December 10 - 16, 2023},
  year         = {2023},
  url          = {http://papers.nips.cc/paper\_files/paper/2023/hash/7a92bcdede88c7afd108072faf5485c8-Abstract-Datasets\_and\_Benchmarks.html},
  bibsource    = {dblp computer science bibliography, https://dblp.org}
}

@Misc{seipp-et-al-zenodo2022,
  author =       "Jendrik Seipp and {\'A}lvaro Torralba and J{\"o}rg Hoffmann",
  title =        "{PDDL} Generators",
  publisher =    "Zenodo",
  year =         "2022",
  howpublished = "\url{https://doi.org/10.5281/zenodo.6382173}"
}
\appendix

\section{Benchmark Scope and Generality}
\label{sec:appendix_benchmark_scope}

Our conclusions are derived from ACPBench-Hard, but ACPBench-Hard should not be viewed as an arbitrary natural-language benchmark. Like other prominent LLM planning benchmarks such as PlanBench \citep{DBLP:conf/nips/ValmeekamMHSK23} and AutoPlanBench \citep{stein2025automating}, it is built on a PDDL/IPC-style symbolic planning problem generator \citep{seipp-et-al-zenodo2022}; these benchmarks differ primarily in the scaffolding layer through which formal planning instances are exposed to LLMs, rather than in the underlying notion of planning itself. ACPBench-Hard is the most recent, and is especially suitable for our analysis because it provides both domain diversity and task diversity: it covers 13 PDDL domains and exposes them through eight open-ended reasoning tasks spanning action-level, state-level, and plan-level queries, rather than only through end-to-end plan generation. We therefore interpret our results as evidence about latent competencies in PDDL-based symbolic planning evaluation. Additional benchmarks would be valuable as scaffold-robustness checks, but they would primarily test robustness to alternative LLM-facing interfaces.

\section{On the Label \emph{Structural Enumeration}}
In classical AI, planning is synonymous with search over a state space \citep{DBLP:journals/ai/BonetG01}. If we were to label the second dimension \emph{search complexity}, it would misleadingly suggest that it captures all of planning-as-search, and that operational reasoning lies outside the scope of search. We therefore adopt the term structural enumeration to denote specifically the ability to engage with the combinatorial exploration of the state space, and clarifies that it is one facet of planning-as-search, not a proxy for the entire activity.

\section{Cross-Family Competence Trajectories}
\label{sec:appendix_cross_family}

\begin{figure}[ht]
  \centering
  \includegraphics[width=\columnwidth]{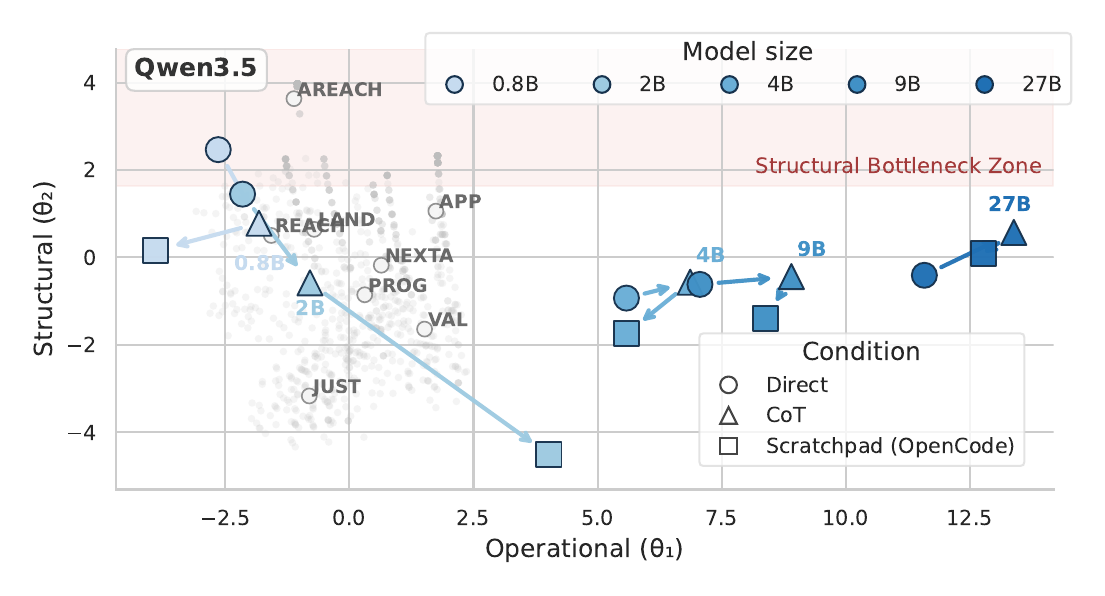}
  \caption{Asymmetric scaling of planning competence. Pale points: item difficulties on the two latent dims. Colored trajectories: ability estimates under Direct $\rightarrow$ CoT $\rightarrow$ Scratchpad. Operational ability (Dim~1) increases with scale ($\beta_{\log\text{size}}=4.34$) and prompting (CoT $1.81$, Scratchpad $1.97$); structural enumeration (Dim~2) does not ($+0.07$, $-0.39$, $-1.19$). The shaded zone marks the top quartile of structural difficulty, where scale provides little gain.}
  \label{fig:qwen3_teaser_image}
\end{figure}

\begin{figure}[ht]
  \centering
  \includegraphics[width=\columnwidth]{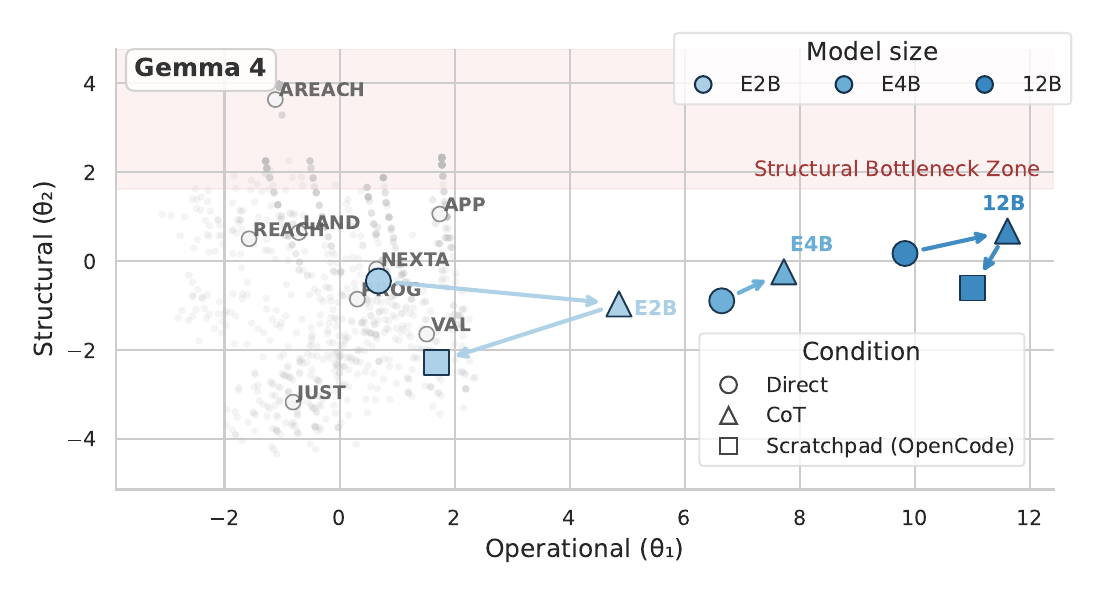}
  \caption{Gemma~4 competence trajectories under joint calibration.
  Pale points show shared item difficulty coordinates; colored points show
  Gemma~4 ability estimates for Direct, CoT, and Scratchpad conditions
  (Direct and CoT only for E4B, whose scratchpad run is excluded).
  The dominant movement is rightward along the operational axis; structural
  ability remains comparatively flat.
  The shaded band marks the top quartile of structural difficulty
  ($b_2 \geq 1.63$), where all conditions remain bottlenecked.}
  \label{fig:appendix:gemma4_teaser_image}
\end{figure}

\begin{figure}[ht]
  \centering
  \includegraphics[width=\columnwidth]{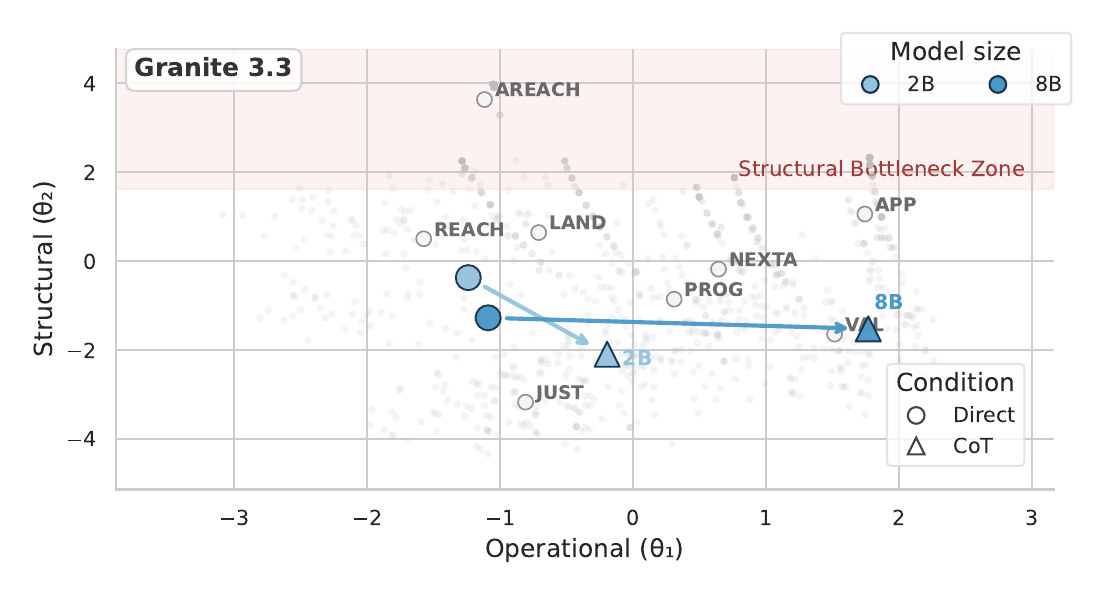}
  \caption{Granite~3.3 competence trajectories under joint calibration.
  Pale points show shared item difficulty coordinates; colored points show
  Granite~3.3 ability estimates for Direct and CoT
  conditions (both scratchpad runs are excluded).
  As with Qwen and Gemma, the primary movement is rightward along the
  operational axis, with limited structural gains.}
  \label{fig:appendix:granite_teaser_image}
\end{figure}

\Cref{sec:asymmetric_scaling} of the main text reports that the
asymmetric scaling pattern, i.e., operational ability improving with scale and
test-time reasoning mechanisms while structural enumeration remains comparatively flat, is
consistent across model families under joint calibration.
\Cref{fig:qwen3_teaser_image} shows that larger Qwen3.5 models move primarily along the operational axis, without a comparable rise in structural ability. The Gemma~4 trajectories in \Cref{fig:appendix:gemma4_teaser_image} show the same scale-related asymmetry. In both families, scratchpad can lower structural ability relative to CoT. \Cref{fig:appendix:granite_teaser_image} shows that CoT increases operational ability at both Granite~3.3 sizes while structural ability decreases. These comparisons use only Direct and CoT for Granite~3.3 and omit scratchpad for Gemma~4-E4B.

All three figures share the same jointly calibrated coordinate system:
item difficulties $(b_1,b_2)$ are estimated once from the stacked
response matrix spanning all families, and respondent abilities
$(\theta_1,\theta_2)$ are placed in that common space.
This ensures that horizontal movement corresponds to gains in operational
reasoning and vertical movement to gains in structural enumeration across
all families.

\subsection{Summary of Cross-Family Evidence}

Across Qwen3.5 (dense 0.8B--27B sweep) and Qwen3.8 (27B), Gemma~4 (E2B, E4B, 12B), and
Granite~3.3 (2B, 8B), the same qualitative pattern holds under joint
calibration: model scale and reasoning-style prompting primarily
improve operational ability, while structural enumeration remains a
persistent bottleneck.
The absolute ability levels differ across families, but the direction
of improvement is stable.
This cross-family consistency suggests that the two-dimensional
noncompensatory structure and its asymmetric response to scale and
prompting are not artifacts of a single model series, but a more
general property of the LLM planning behavior captured by ACPBench-Hard.

\section{Inference Condition System Prompts}
\label{sec:appendix_prompts}

This appendix documents the exact system prompts and generation
configurations for the three core inference conditions used in the main
text: \texttt{no\_think} (Direct), \texttt{think} (Chain-of-Thought),
and \texttt{opencode\_agent} (Scratchpad in the OpenCode agent harness).  The Direct and CoT prompts are defined in
\texttt{src/acpbench\_evaluation/config.py}; per-model sampling is declared in the campaign specification (\Cref{tab:gen_config}).

\subsection{Direct (\texttt{no\_think})}

\paragraph{Generation configuration.}~\\ 

Instruct sampling profile of each model (\Cref{tab:gen_config}), with
\texttt{enable\_thinking=False} and no tools.

\paragraph{System prompt.}~\\ 

\begin{lstlisting}[
    breaklines=true,
    breakatwhitespace=false,
    basicstyle=\small\ttfamily,
    columns=fullflexible,
    frame=single,
    aboveskip=8pt,
    belowskip=8pt,
]
You are an expert in classical planning and reasoning about actions,
state transitions, and goals. Answer the question precisely following
the specified output format. In think mode, keep reasoning finite and
search-oriented: enumerate candidates, check them, and stop once the
answer is determined. Wrap the reasoning trace in <think>...</think>.
Be concise and avoid repetition.
\end{lstlisting}

The system prompt is followed by zero to two few-shot examples showing
question\(\to\)answer pairs (without reasoning traces) and the
task-specific format instruction (see \Cref{sec:appendix_format_instructions}).
The test question is appended as a user message.

\subsection{Chain-of-Thought (\texttt{think})}

\paragraph{Generation configuration.}~\\ 

Thinking sampling profile of each model (\Cref{tab:gen_config}), with
thinking enabled and no tools, followed by an answer-extraction turn.

\paragraph{System prompt.}
Identical to the Direct condition system prompt shown above.

\paragraph{Few-shot example structure.}
The few-shot examples for \texttt{think} include explicit
\texttt{<think>...</think>} reasoning traces:

\begin{lstlisting}[
    breaklines=true,
    breakatwhitespace=false,
    basicstyle=\small\ttfamily,
    columns=fullflexible,
    frame=single,
    aboveskip=8pt,
    belowskip=8pt,
]
Example 1:
<context>...</context>
<question>...</question>
<format_instruction>

<think>
[step-by-step reasoning: enumerate candidates, check preconditions,
verify effects, determine answer]
</think>

[final answer]
\end{lstlisting}

\subsection{Scratchpad in the OpenCode Agent Harness
(\texttt{opencode\_agent})}

\paragraph{Generation configuration.}~\\ 
Thinking sampling profile of each model (\Cref{tab:gen_config}), 32{,}768 output tokens per turn, and a 20-minute limit per item. Each item runs in a fresh OpenCode workspace outside the repository; permissions allow only workspace file reads and edits.

\paragraph{Task prompt.}~\\ 
Each workspace contains a \texttt{TASK.md} with the item context, question, and format instruction, preceded by the following scaffold (abridged):

\begin{lstlisting}[
    breaklines=true,
    breakatwhitespace=false,
    basicstyle=\small\ttfamily,
    columns=fullflexible,
    frame=single,
    aboveskip=8pt,
    belowskip=8pt,
]
You are solving one ACPBench planning item. Use the OpenCode workspace
as an external scratchpad for structural enumeration, not as a place
for free-form rambling.

Use only workspace file reads and file writes/edits. Before writing the
final answer, create and maintain these files in the sample workspace:
- frontier.md: reachable state groups, candidate transitions, and the
  next state groups to inspect.
- visited.md: state groups, facts, actions, or landmarks already checked.
- dead_ends.md: candidates ruled out, with the exact reason.
- candidate_answers.md: every candidate final answer considered and
  whether it is valid.
- final_answer.txt: the final answer only, in the exact task format.

For structural tasks such as land, reach, and areach, prioritize explicit
reachability, landmark, and necessary-subgoal checks. For operational
controls such as app, val, and prog, still use the files above, but keep
the search local and finite.

Stop when the answer is determined.
\end{lstlisting}

The content of \texttt{final\_answer.txt} is scored directly by the benchmark validators.

\section{Task Format Instructions}
\label{sec:appendix_format_instructions}

Every task-specific user prompt appends a short format instruction after
the question text.  The full set of instructions is defined in
\texttt{src/acpbench\_evaluation/config.py}
(\texttt{TASK\_FORMAT\_INSTRUCTIONS}).

\paragraph{Generative-format instructions.}
\Cref{tab:format_instructions} specifies the required answer form for each generative task. For example, action answers use parenthesized expressions, while \texttt{prog} requires two lists of ground propositions. These instructions constrain the output representation without supplying candidate answers.

\begin{table*}[h]
\centering
\scriptsize
\caption{Task format instructions for generative (free-text) questions.}
\label{tab:format_instructions}
\begin{tabular}{ll}
\toprule
Task & Format instruction \\
\midrule
\texttt{app} &
Each action starts with an opening parenthesis and ends with closing
parenthesis. Provide only the actions. \\
\texttt{areach} &
Each action starts with an opening parenthesis and ends with closing
parenthesis. Provide one action or None. \\
\texttt{just} & (none) \\
\texttt{land} &
Provide only the ground proposition or None. \\
\texttt{nexta} &
Each action starts with an opening parenthesis and ends with closing
parenthesis. Provide only the action. \\
\texttt{prog} &
Provide only the two lists with the ground propositions. \\
\texttt{reach} &
Provide one proposition or None. \\
\texttt{val} &
Provide only the index of the action. \\
\bottomrule
\end{tabular}
\end{table*}

\paragraph{Boolean-format instructions.}
All eight tasks use the same instruction: ``Answer only yes or no.''

\paragraph{MCQ-format instructions.}
All eight tasks use the same instruction: ``Answer only the single
correct option label (A, B, C, or D).''

\section{Inference Infrastructure and Model Serving}
\label{sec:appendix_infrastructure}

All models were served with \textsc{vllm} \citep{kwon2023efficient} using an OpenAI-compatible chat-completion endpoint. The server ran on a single node with four NVIDIA RTX 5090 GPUs (32\,GB each), and we enabled prefix caching to accelerate shared-prompt evaluations. Generation parameters were controlled via the chat-completion API; each model uses the sampling settings recommended on its official model card, listed in \Cref{tab:gen_config}.

\begin{table}[t]
\centering
\scriptsize
\setlength{\tabcolsep}{3pt}
\caption{Per-model sampling. Direct uses the instruct profile; CoT and Scratchpad use the thinking profile. Values are temperature / top-$p$ / top-$k$ / presence penalty; the repetition penalty is 1.0 throughout. The cap is the first-turn token limit of the thinking profile.}
\label{tab:gen_config}
\resizebox{\columnwidth}{!}{%
\begin{tabular}{lllr}
\toprule
Models & Instruct (Direct) & Thinking (CoT, Scratchpad) & Cap \\
\midrule
Qwen3.5 0.8B, 2B & 1.0 / 1.0 / 20 / 2.0 & 1.0 / 0.95 / 20 / 1.5 & 32{,}768 \\
Qwen3.5 4B--27B  & 0.7 / 0.8 / 20 / 1.5 & 1.0 / 0.95 / 20 / 1.5 & 32{,}768 \\
Qwen3.8 27B\textsuperscript{a} & 0.7 / 0.8 / 20 / 1.5 & 1.0 / 0.95 / 20 / 0.0 & 16{,}384 \\
Gemma~4 E2B, E4B, 12B & 1.0 / 0.95 / 64 / 0.0 & 1.0 / 0.95 / 64 / 0.0 & 16{,}384 \\
Granite~3.3 2B, 8B & 0.6 / 0.9 / 50 / 0.0 & 0.6 / 0.9 / 50 / 0.0 & 8{,}192 \\
\bottomrule
\multicolumn{4}{l}{\textsuperscript{a} Reasoning effort \texttt{medium}, thinking budget 16{,}384 tokens.}
\end{tabular}%
}
\end{table}

\paragraph{Key design choices.}
\textbf{Sampling.}
Each model uses its official recommended sampling rather than greedy decoding, so that every family is evaluated in the regime its developers tuned; each item--condition pair is sampled once.

\textbf{Answer extraction.}
CoT responses are followed by a short answer-extraction turn with thinking disabled, which pulls the final answer out of the reasoning trace before scoring; Scratchpad answers are read from \texttt{final\_answer.txt}.

\textbf{Tool use.}
Scratchpad runs use OpenCode's tool loop with only file read and edit permissions, terminating when the agent stops or the 20-minute item limit is reached.

All model-specific vLLM settings (tensor-parallel size, GPU memory fraction, family-specific parsers) and the pipeline scripts (\texttt{scripts/pipeline/collect\_model.py}, \texttt{scripts/pipeline/evaluate.py}) are available in the accompanying code repository.

\section{Semantic Codebook}
\label{sec:appendix_semantic_codebook}

This appendix documents the full semantic tagging codebook used to assign interpretable labels to ACPBench tasks.
Each task receives a set of task-default binary tags drawn from three complementary categories: \textit{Reasoning-type}, which captures the cognitive operation the model must perform; \textit{Output-demand}, which captures how the answer must be expressed; and \textit{Structural-demand}, which captures the computational structure that makes the problem difficult.
These tags were specified before any IRT model fit and are intended as transparent task-default semantic annotations rather than manually reviewed item-by-item labels.

The main text does not use this codebook in its interpretation of the latent dimensions; the interpretation in \Cref{sec:dimension_interpretation} is grounded in behavioral, representational, and intervention evidence. The codebook is reported here for transparency and auditability, together with its correlation analysis (\Cref{sec:appendix_tag_correlations}), which is consistent with the interpretation but is not needed to establish it.

\subsection{Tag Glossary}

\begin{table*}[h]
\centering
\scriptsize
\caption{Semantic tag definitions with their categories.}
\label{tab:semantic_tag_glossary}
\begin{tabular}{@{}l p{0.55\textwidth} l@{}}
\toprule
\textbf{Tag} & \textbf{Definition} & \textbf{Category} \\
\midrule
local\_state\_check & Reasoning about the truth of facts in a single state & Reasoning-type \\
state\_transition & Computing the effects of applying an action & Reasoning-type \\
plan\_validation & Auditing a proposed plan for correctness & Reasoning-type \\
plan\_justification & Determining whether plan steps are necessary & Reasoning-type \\
global\_reachability & Reasoning about whether facts can hold in any reachable state & Reasoning-type \\
action\_reachability & Reasoning about whether an action can ever become applicable & Reasoning-type \\
landmark\_reasoning & Identifying facts that must hold in all valid plans & Reasoning-type \\
next\_action\_construction & Constructing a goal-directed next action & Reasoning-type \\
\addlinespace
verification\_output & Answer requires verifying a given claim & Output-demand \\
construction\_output & Answer requires constructing a novel answer & Output-demand \\
single\_witness\_output & Answer is a single fact, action, or index & Output-demand \\
set\_enumeration\_output & Answer is a set or list & Output-demand \\
negative\_unreachability\_case & Answer requires certifying that no valid witness exists for a reachability-style query & Output-demand \\
\addlinespace
requires\_search & Requires exploring the state space & Structural-demand \\
requires\_composition\_over\_steps & Requires composing across multiple plan steps & Structural-demand \\
requires\_counterfactual\_consistency & Requires consistent reasoning about counterfactuals & Structural-demand \\
requires\_goal\_structure\_tracking & Requires tracking the relationship of facts to goals & Structural-demand \\
\bottomrule
\end{tabular}
\end{table*}

The 17 tags enumerated in \Cref{tab:semantic_tag_glossary} span three complementary semantic families.
Reasoning-type tags capture \textit{what} cognitive operation the model must perform, for example checking local state facts versus reasoning about global reachability.
Output-demand tags capture \textit{how} the answer must be expressed, for example verification versus construction, single-witness reporting versus set enumeration, and negative unreachability certification.
Structural-demand tags capture the computational properties that make a problem difficult, for example search, multi-step composition, counterfactual consistency, and goal-structure tracking.

\subsection{Task-to-Tag Mapping}

Each ACPBench task is assigned a fixed set of semantic tags based on its canonical definition.
The mapping is shown in \Cref{tab:task_tag_mapping}.

\begin{table*}[ht]
\centering
\setlength{\tabcolsep}{3pt}
\scriptsize
\caption{Task-to-tag mapping for all eight ACPBench tasks. Tags are assigned at the task-default level.}
\label{tab:task_tag_mapping}
\begin{tabularx}{\textwidth}{@{}l>{\raggedright\arraybackslash}p{0.25\textwidth}>{\raggedright\arraybackslash}X@{}}
\toprule
\textbf{Task} & \textbf{Description} & \textbf{Active Tags} \\
\midrule
app & Applicable-action enumeration & local\_state\_check, state\_transition, verification\_output, set\_enumeration\_output \\
prog & Apply local effects and report delta & local\_state\_check, state\_transition, verification\_output \\
val & Audit plan, find first invalid step & plan\_validation, verification\_output, requires\_composition\_over\_steps, requires\_counterfactual\_consistency \\
just & Validate necessary steps, simplify a plan & plan\_justification, verification\_output, requires\_composition\_over\_steps, requires\_counterfactual\_consistency \\
reach & Which facts can hold in reachable states & global\_reachability, single\_witness\_output, negative\_unreachability\_case, requires\_search, requires\_goal\_structure\_tracking \\
areach & Can an action ever become applicable & action\_reachability, single\_witness\_output, negative\_unreachability\_case, requires\_search, requires\_goal\_structure\_tracking \\
land & Construct a non-trivial landmark & landmark\_reasoning, construction\_output, single\_witness\_output, requires\_search, requires\_goal\_structure\_tracking \\
nexta & Construct a goal-directed next action & next\_action\_construction, construction\_output, single\_witness\_output, requires\_search, requires\_composition\_over\_steps \\
\bottomrule
\end{tabularx}
\end{table*}

Tags are assigned at the task-default level, which means each item inherits the semantic tag pattern of its parent task.
This codebook was specified before any IRT model fit was conducted, ensuring that the semantic dimensions used for interpretation are not confounded by post-hoc knowledge of model behavior.

\subsection{Tag--Dimension Correlations}
\label{sec:appendix_tag_correlations}

Because the tags are assigned a priori by the authors, we treat these correlations as secondary evidence: they are consistent with the interpretation in \Cref{sec:dimension_interpretation}, but they cannot independently establish it.

\begin{table}[t]
\centering
\setlength{\tabcolsep}{3pt}
\scriptsize
\caption{Selected direct correlations between task-default semantic tags and IRT difficulty dimensions. All reported associations survive Benjamini--Hochberg correction (see \Cref{sec:appendix_robustness})}
\label{tab:tag_correlations}
\begin{tabularx}{\columnwidth}{@{}>{\raggedright\arraybackslash}Xlrr@{}}
\toprule
\textbf{Tag} & \textbf{Dim.} & \textbf{$r$} & \textbf{$p$-value} \\
\midrule
\path{requires_counterfactual_consistency} & $b_2$ & $-$0.641 & 3.56e-121 \\
\path{verification_output} & $b_2$ & $-$0.504 & 4.41e-68 \\
\path{requires_search} & $b_2$ & +0.504 & 4.41e-68 \\
\path{requires_goal_structure_tracking} & $b_2$ & +0.534 & 8.39e-78 \\
\path{requires_composition_over_steps} & $b_2$ & $-$0.587 & 3.53e-97 \\
\path{requires_goal_structure_tracking} & $b_1$ & $-$0.705 & 3.19e-157 \\
\bottomrule
\end{tabularx}
\end{table}

The structural axis ($b_2$) shows symmetric opposite associations with \path{verification_output} ($r = -0.504$) and \path{requires_search} ($r = +0.504$). On the operational axis ($b_1$), \path{requires_goal_structure_tracking} correlates negatively ($r = -0.705$). These alignments are directionally consistent with the interpretation in \Cref{sec:dimension_interpretation}, but they are not needed to establish it.

\section{PDDL-Derived Structural Properties}
\label{sec:appendix_pddl_properties}
This appendix reports the correlations between four PDDL-derived properties and the two IRT difficulty dimensions ($b_1$, $b_2$), computed over the 1,032 items whose PDDL problem could be parsed (981 with a computed plan length). All properties are computed from grounded PDDL representations independently of any LLM responses. These are observational correlations, not causal evidence.

As \Cref{tab:pddl_property_correlations} shows, the branching proxy and the predicate count correlate positively with $b_2$ ($r=0.156$ and $r=0.118$) and negligibly with $b_1$ ($r=-0.014$ and $r=0.019$). Plan length shows the largest association, negative on $b_2$ ($r=-0.279$) and weakly positive on $b_1$ ($r=0.112$). The action-count correlations are near zero for both dimensions.

\subsection{Full Correlation Table}

\begin{table}[h!]
\centering
\scriptsize
\caption{Spearman correlations between PDDL properties and IRT difficulty dimensions ($b_1$, $b_2$), with $p$-values in parentheses. $n=1032$ items ($n=981$ for plan length).}
\label{tab:pddl_property_correlations}
\begin{tabular}{lrr}
\toprule
\textbf{Property} & \textbf{$b_1$ ($r$)} & \textbf{$b_2$ ($r$)} \\
\midrule
\quad branching\_proxy & $-$0.014 (0.66) & 0.156 (4.7e-07) \\
\quad num\_predicates & 0.019 (0.54) & 0.118 (1.4e-04) \\
\quad num\_actions & 0.000 (0.99) & 0.020 (0.53) \\
\quad plan\_length & 0.112 (4.5e-04) & $-$0.279 ($<$1e-08) \\
\bottomrule
\end{tabular}
\end{table}

These properties are parser-backed from PDDL specifications. The branching proxy and predicate count show modest positive correlations with $b_2$ and none with $b_1$, consistent with the interpretation that $b_2$ captures structural complexity. Plan length does not follow this pattern: items with longer plans are easier on $b_2$, so problem size alone does not account for the second dimension.

\section{Scaling Regression}
\label{sec:appendix_scaling_regression}

To quantify the trajectory pattern in \Cref{fig:all_family_trajectories}, we fit a simple linear regression to the jointly calibrated ability estimates. Each observation is a model--condition pair. The dependent variable is the estimated ability on one latent dimension, \(\theta_{id}\), and the predictors are centered log model size and inference-condition indicators:
\begin{equation}
\begin{aligned}
  \theta_{id} = \; & \alpha_d 
  + \beta_{d,\mathrm{size}} \log N_i \\
  & + \beta_{d,\mathrm{CoT}} \,\mathbb{I}[\mathrm{CoT}] \\
  & + \beta_{d,\mathrm{Scratchpad}} \,\mathbb{I}[\mathrm{Scratchpad}] 
  + \epsilon_{id}
\end{aligned}
\label{eq:ability_regression}
\end{equation}
where \(d \in \{1,2\}\), \(N_i\) is the parameter count of model \(i\), and Direct inference is the reference condition. We fit the regression separately for the operational dimension \((\theta_1)\) and the structural dimension \((\theta_2)\).

\begin{table}[h]
\centering
\small
\setlength{\tabcolsep}{3pt}
\caption{Regression of jointly calibrated ability estimates on model size
and inference condition. Operational ability increases with model size and
reasoning-style prompting, while structural enumeration shows no comparable
positive effect. Direct inference is the reference condition.}
\label{tab:scaling_regression}
\begin{tabular}{lrr}
\toprule
Predictor & Operational \(\theta_1\) & Structural \(\theta_2\) \\
\midrule
Centered log model size & \(+4.34\) & \(+0.07\) \\
CoT & \(+1.81\) & \(-0.39\) \\
Scratchpad & \(+1.97\) & \(-1.19\) \\
\bottomrule
\end{tabular}
\end{table}

The point estimates in \Cref{tab:scaling_regression} quantify the asymmetry in \Cref{fig:all_family_trajectories}. The log-size coefficient is $+4.34$ for operational ability and $+0.07$ for structural ability. Relative to Direct, the CoT and scratchpad coefficients are positive on the operational dimension ($+1.81$ and $+1.97$) and negative on the structural dimension ($-0.39$ and $-1.19$). The fitted pattern therefore concentrates gains along the operational axis. These pooled associations across model families and inference conditions are not causal estimates of the effects of adding parameters or computation.

\section{IRT Model Diagnostics}
\label{sec:appendix_irt_diagnostics}

\subsection{Task (Item) Difficulty Structure}
\begin{table}[h]
\centering
\scriptsize
\caption{Task-level mean difficulties in the 2D noncompensatory model. Positive values indicate harder-than-average on that dimension. Tasks separate into two groups: those requiring local transition reasoning (Dim1) and those requiring global state-space reasoning (Dim2).}
\label{tab:task_anchors}
\begin{tabular}{lrr}
\toprule
Task & $b_1$ (Dim1) & $b_2$ (Dim2) \\
\midrule
\texttt{app}    & +1.74 & +1.06 \\
\texttt{val}    & +1.51 & -1.64 \\
\texttt{nexta}  & +0.64 & -0.18 \\
\texttt{prog}   & +0.31 & -0.85 \\
\texttt{land}   & -0.71 & +0.64 \\
\texttt{just}   & -0.81 & -3.17 \\
\texttt{areach} & -1.12 & +3.63 \\
\texttt{reach}  & -1.57 & +0.50 \\
\bottomrule
\end{tabular}
\end{table}

\begin{quote}
\small
\textit{Note:} \texttt{areach} is extremely hard on Dim2 only; \texttt{just} is easiest overall (both dimensions easy). Dim1 correlates with planning/local transitions; Dim2 with global state-space reasoning.
\end{quote}

Key observations:

\begin{itemize}
  \item \textbf{app} is the hardest Dim1 (general planning) challenge. It anchors the rotation: the model constrains app's raw task effect on the final dimension to zero, which fixes the rotational ambiguity of the 2D solution (the reported, mean-centred $b_2$ of app is therefore not zero).
  \item \textbf{areach} has an extreme Dim2 difficulty ($b_2$ = 3.63), placing it at the floor for virtually all model-condition pairs. It provides little discriminative power for Dim2 ability estimation.
  \item \textbf{just} and \textbf{reach/land} form opposite poles: just is near-trivial, while reach and land require structural traversal reasoning.
  \item Tasks splitting across dimensions (land, areach) validate the noncompensatory structure: you cannot substitute excess Dim1 ability for Dim2 deficits.
\end{itemize}

\subsection{IRT Training Configuration}

All IRT models are trained via maximum a posteriori~(MAP) estimation with the following priors and constraints:

\begin{itemize}
    \item \textbf{Gaussian priors:} $\sigma_u=2.0$ for respondent ability residuals, $\sigma_\alpha=1.0$ for task effect parameters, $\sigma_\varepsilon=1.0$ for item residual parameters. Regression coefficients receive a weak prior with $\sigma_B=10.0$.
    \item \textbf{Anchor constraint:} The \texttt{app} task~(index~0) is constrained to have zero difficulty on the final latent dimension, enforced with a penalty weight of $100.0$.
    \item \textbf{Optimizer:} full-batch L-BFGS in double precision, run until the largest absolute gradient falls below $10^{-3}$; all reported fits converged.
    \item \textbf{Random starts:} 8 random initializations for noncompensatory models, whose objective has several modes, keeping the best MAP objective; 1 for compensatory models.
    \item \textbf{Cross-validation:} 5-fold cross-validation with random respondent-by-item cell holdout, stratified by item to ensure each item appears in every fold.
    \item \textbf{Latent regression specification:}
    \begin{equation*}
        \theta_{id} = \mathbf{B}_d^{\top} \mathbf{x}_i + u_{id}
    \end{equation*}
    where $\mathbf{x}_i$ includes centered log model size, model family indicators (Qwen3.8, Gemma~4, Granite~3.3; Qwen3.5 as reference), and inference condition indicators (CoT, Scratchpad; Direct as baseline), and $u_{id}$ is a respondent-specific residual.
\end{itemize}

\subsection{Activation-Based Probing of Latent Difficulty}
\label{sec:appendix_probe}

To examine whether the two latent difficulty dimensions are reflected in model representations, we trained linear ridge regression probes to predict item-level IRT difficulty coordinates $(b_1, b_2)$ from the hidden states of Qwen~3.5 models. The probes were evaluated under leave-one-task-out cross-validation: for each fold, all items belonging to one task were held out, ensuring that the probe cannot exploit task identity alone.

We compared the hidden-state probe against four baselines:
\begin{itemize}
    \item \textbf{Task-ID baseline:} an 8-dimensional one-hot task identity vector. If hidden states only encoded task membership, they would not outperform this baseline.
    \item \textbf{PDDL-syntax baseline:} a regressor trained solely on surface PDDL features (branching proxy, action count, predicate count, plan length). This tests whether the difficulty coordinates merely reflect superficial problem complexity.
    \item \textbf{PDDL-residualised hidden states:} we first removed the component of hidden-state variance linearly predictable from PDDL features, then trained the probe on the residual. This assesses whether neural representations carry planning-relevant information beyond surface syntax.
    \item \textbf{Shuffled-target control:} the $(b_1, b_2)$ targets were randomly permuted before training, keeping the hidden states unchanged. This confirms that the probe is not fitting noise.
\end{itemize}

The experiment focused on Qwen~3.5~9B and Qwen~3.5~27B, both under the chain-of-thought (CoT) condition. For each model we swept all transformer layers and selected the layer minimizing joint mean squared error (MSE) on the held-out tasks.

\paragraph{Key findings.}
\begin{itemize}
    \item The best-predicting layer shifts deeper with scale: layer~13 for the 9B model, layer~52 for the 27B model.
    \item Hidden-state probes outperform all baselines in joint MSE for both model sizes (Table~\ref{tab:probe_mse}).
    \item The predictive signal strengthens with model size: 27B's joint MSE (2.72) is $\sim$17\% lower than 9B's (3.29).
\end{itemize}

\begin{table}[h]
\centering
\scriptsize
\caption{Joint MSE of latent-difficulty probes under leave-one-task-out cross-validation. Lower is better. The hidden-state ridge probe outperforms all baselines.}
\label{tab:probe_mse}
\begin{tabular}{lcc}
\toprule
Probe & 9B Joint MSE $\downarrow$ & 27B Joint MSE $\downarrow$ \\
\midrule
Hidden ridge (best layer) & 3.29 & 2.72 \\
Task-ID baseline           & 4.00 & 4.00 \\
PDDL/proxy-only baseline   & 3.81 & 3.81 \\
PDDL-residualised hidden   & 3.37 & 3.60 \\
Shuffled-target control    & 3.99 & 4.24 \\
\bottomrule
\end{tabular}
\end{table}

These results show that the two IRT difficulty dimensions can be linearly decoded from internal activations beyond what task identity or surface PDDL properties explain. The deepening and strengthening of the signal with scale is consistent with larger models encoding planning difficulty more explicitly, providing representational-level convergent evidence for the two-dimensional structure.

\subsection{Selective Neuron Analysis}
\label{sec:appendix_sna}

To investigate whether the latent difficulty signal is neurally diffuse or concentrated in distinct subpopulations, we conducted a selective neuron analysis on the best-predicting layer of the strongest probe model (Qwen~3.5~27B, layer~52). For all 5,120 neurons in the feed-forward output of that layer, we computed Pearson correlations with $b_1$ and $b_2$ across all items. We applied Benjamini-Hochberg FDR correction ($\alpha = 0.05$) and classified a neuron as selective for one dimension if its absolute correlation with that dimension was larger than with the other \emph{and} FDR-significant. We then trained ridge probes to predict $b_1$ and $b_2$ using only the top-50 most selective neurons for each dimension, under the same leave-one-task-out folds, re-selecting the neurons inside each training fold so that the held-out task never informs the selection.

\paragraph{Findings.}
\begin{itemize}
    \item Selectivity is widespread: 2,643 neurons are $b_1$-selective and 2,317 are $b_2$-selective under this criterion.
    \item The top-50 selective sets for the two dimensions are disjoint, indicating that the strongest neural correlates of operational and structural reasoning occupy distinct subspaces.
    \item Probes built from these small neuron subsets do not outperform the full-layer probe on their respective targets (Table~\ref{tab:sna_mse}): the difficulty signal is not concentrated in a sparse subset beyond what the full layer already captures.
\end{itemize}

\begin{table}[h]
\centering
\scriptsize
\caption{MSE of top-50 selective neuron probes vs.\ full-layer probe on the same target (Qwen~3.5~27B, layer~52), with neuron selection repeated inside each training fold. The full-layer probe achieves lower error on both targets.}
\label{tab:sna_mse}
\begin{tabular}{lcc}
\toprule
Target & Top-50 Subset MSE & Full-layer MSE \\
\midrule
$b_1$ & 1.53 & 1.25 \\
$b_2$ & 4.75 & 4.18 \\
\bottomrule
\end{tabular}
\end{table}

The disjoint top-50 sets provide neural-level convergent evidence that the two latent competencies correspond to partially separable activation subspaces, aligning with the noncompensatory functional form in which weakness in one dimension cannot be offset by strength in the other. Note that this analysis is correlational and is offered as exploratory representational evidence.

\subsection{Statistical Robustness Checks}
\label{sec:appendix_robustness}

We performed a suite of statistical controls to verify that the two-dimensional noncompensatory structure is not an artifact of chance task labels, marginal response patterns, or uncorrected multiple testing.

\paragraph{Null permutation tests.} We compared the observed 2D noncompensatory model’s negative log-likelihood (NLL) against two null distributions: (i) 1,000 random permutations of task labels across items, and (ii) 1,000 random shuffles of the correctness labels while preserving item and respondent margins. In both cases, the observed NLL is lower than every permuted or shuffled sample ($p < 0.001$; Table~\ref{tab:null_tests}), ruling out chance label assignments or marginal effects as the source of the structure.

\begin{table}[h]
\centering
\scriptsize
\setlength{\tabcolsep}{3pt}
\caption{Null permutation tests. The observed NLL is lower than all 1,000 resamples for both null models.}
\label{tab:null_tests}
\begin{tabular}{lccc}
\toprule
Null Test & Resamples & Mean NLL & Observed NLL \\
\midrule
Task-label permutation   & 1{,}000 & 0.3175 & 0.2947 \\
Correctness-label shuffle & 1{,}000 & 0.3155 & 0.2947 \\
\bottomrule
\end{tabular}
\end{table}

\paragraph{Task-collapsed refit.} Collapsing the eight task-specific difficulty parameters into a single global task effect (i.e., removing the task-level structure) raises the evaluation NLL by 0.0198 and lowers accuracy (Table~\ref{tab:task_collapsed}). This indicates that task-specific difficulty structure improves model fit beyond a simple item pool, confirming that the task-level grouping carries meaningful information.

\begin{table}[h]
\centering
\scriptsize
\caption{Comparison of task-specific and task-collapsed refits. Removing task structure worsens fit.}
\label{tab:task_collapsed}
\begin{tabular}{lccc}
\toprule
Refit & Eval NLL & Eval Acc. & $\Delta$ NLL \\
\midrule
Task-specific (8 tasks) & 0.2947 & 0.8739 & -- \\
Task-collapsed         & 0.3145 & 0.8718 & +0.0198 \\
\bottomrule
\end{tabular}
\end{table}

\paragraph{FDR correction for semantic codebook.} The appendix-level codebook analysis (\Cref{sec:appendix_semantic_codebook}) reports Spearman correlations between item-level codebook tags and the $b_1$, $b_2$ coordinates. Applying Benjamini-Hochberg correction ($\alpha = 0.05$) to all 34 reported associations retains 30 as significant (Table~\ref{tab:fdr_top} shows the associations selected in \Cref{tab:tag_correlations}). The core semantic pattern—operational reasoning (Dim1) and structural enumeration (Dim2)—remains robust to multiplicity adjustment.

\begin{table}[h]
\centering
\scriptsize
\caption{Selected semantic-tag correlations surviving FDR correction ($q < 0.05$). Full list available in \Cref{sec:appendix_semantic_codebook}.}
\label{tab:fdr_top}
\resizebox{\columnwidth}{!}{%
\begin{tabular}{lcc}
\toprule
Tag & Spearman $r$ & BH $q$ \\
\midrule
\texttt{requires\_goal\_structure\_tracking} ($b_1$)   & $-$0.705 & $1.1 \times 10^{-155}$ \\
\texttt{requires\_counterfactual\_consistency} ($b_2$) & $-$0.641 & $6.0 \times 10^{-120}$ \\
\texttt{requires\_composition\_over\_steps} ($b_2$)    & $-$0.587 & $3.0 \times 10^{-96}$  \\
\texttt{requires\_goal\_structure\_tracking} ($b_2$)   & +0.534 & $2.9 \times 10^{-77}$  \\
\texttt{requires\_search} ($b_2$)                      & +0.504 & $1.1 \times 10^{-67}$  \\
\texttt{verification\_output} ($b_2$)                  & $-$0.504 & $1.1 \times 10^{-67}$  \\
\bottomrule
\end{tabular}%
}
\end{table}

\subsection{Agent Scaffold by Structural Difficulty}
\label{sec:appendix_scaffold}

A natural concern is whether the structural-enumeration bottleneck is an artifact of inference without external state. The Scratchpad condition tests this directly: it runs each item inside the OpenCode agent harness, where the model can externalize its search into workspace files.

On Qwen~3.5~27B (100 items per task, paired across conditions), overall accuracy (Table~\ref{tab:scaffold_overall}) shows that the agent harness improves over Direct (+5.9 points) but remains below CoT (+14.3 points over Direct). More importantly, when we break down accuracy by $b_2$ difficulty quartile (Table~\ref{tab:scaffold_b2}), the hardest structural items ($b_2$ Q4) stay near floor even with the agent harness: accuracy drops from 96.5\% (easiest) to 1.0\% (hardest).

\begin{table}[h]
\centering
\scriptsize
\caption{Overall accuracy of Qwen~3.5~27B under each inference condition on the 800 paired items.}
\label{tab:scaffold_overall}
\begin{tabular}{lcc}
\toprule
Condition & Overall Accuracy & $\Delta$ vs.\ Direct \\
\midrule
Direct                 & 0.478 & -- \\
CoT                    & 0.620 & +0.143 \\
Scratchpad (OpenCode)  & 0.536 & +0.059 \\
\bottomrule
\end{tabular}
\end{table}

\begin{table}[h]
\centering
\scriptsize
\caption{Accuracy of Qwen~3.5~27B by $b_2$ difficulty quartile (fixed item sets, 200 items each). The hardest structural items remain near zero even with the agent harness.}
\label{tab:scaffold_b2}
\begin{tabular}{lccc}
\toprule
$b_2$ Quartile & OpenCode Accuracy & CoT Accuracy & $\Delta$ \\
\midrule
Q1 (low)   & 0.965 & 0.990 & $-$0.025 \\
Q2         & 0.785 & 0.890 & $-$0.105 \\
Q3         & 0.385 & 0.545 & $-$0.160 \\
Q4 (high)  & 0.010 & 0.055 & $-$0.045 \\
\bottomrule
\end{tabular}
\end{table}

The structural-enumeration bottleneck persists under the evaluated OpenCode harness with file-based scratchpad tools, so it is not confined to inference without external state. This result is specific to the tested model, item sets, and tool permissions; it does not establish that every agent harness or form of tool use must fail. The hardest quartile also has little outcome variance, so near-floor accuracy limits what can be inferred about incremental structural gains. The earlier stronger-versus-lightweight scaffold comparison is reported separately in \Cref{sec:appendix_rebuttal_scaffold}.

\subsection{Qualitative Failure Case: Structural Enumeration}
\label{sec:appendix_qual_example}

To complement the quantitative analyses, we retain the logistics-domain OpenCode failure case reported during the rebuttal. It comes from the earlier agent trajectory, not from the newer 800-item collection in \Cref{sec:appendix_scaffold}.

\paragraph{Task.} In a logistics domain, the model must identify a grounded action that can never become applicable in any reachable state. The instance contains three airport locations (\texttt{l0-0}, \texttt{l1-0}, \texttt{l2-0}) and several non-airport locations (e.g., \texttt{l2-2}). The gold answer is \texttt{(unload-airplane p2 a0 l2-2)}, which is impossible because the airplane can never be at \texttt{l2-2}.

\paragraph{Model output.} The model outputs \texttt{(fly-airplane a0 l1-0 l0-0)}. This action is immediately applicable in the initial state, so it is not impossible.

\paragraph{Failure mechanism.} The scratchpad trace reveals a type-structure / closed-world inference error: the model correctly states that a \texttt{fly} action requires both endpoints to be airports, but it infers the set of airports from the initial airplane location rather than from the typed object declarations. Under that mistaken world, it treats \texttt{l0-0} as a non-airport and concludes the \texttt{fly} action is impossible, without verifying whether the action is already applicable. The trace shows an unresolved warning (\textit{“Check if any other location could be an airport”}) that was never closed before finalization. This exemplifies the difficulty of maintaining and reasoning over the complete type structure of a planning problem, a core aspect of the Dim2 structural-enumeration competency.

\section{Intervention Experimental Details}
\label{sec:appendix_interventions}

This appendix documents the design of the intervention experiments in \Cref{sec:interventions}; the design was pre-registered before the intervention runs began.

\subsection{The Oracle Ladder}
\label{sec:appendix_ladder}

The ladder reuses the existing 1{,}040-item ACPBench-Hard response matrices for L1--L2. L3 (CTE) runs on the full item bank (all eight tasks). L4 tiers and distractor controls run on the structural subset (\texttt{reach}, \texttt{areach}, \texttt{land}; 390 items) together with a matched-size sample of operational controls (\texttt{app}, \texttt{prog}, \texttt{val}) under L4-T2, guarding the claim that injection effects are dimension-specific. All new runs use Qwen3.5-27B with $n = 1$ generation per item--condition at the paper's decoding configuration (temperature $0$), matching the configuration of the reused matrices; scoring reuses the benchmark's symbolic validators. The Gemma-4-12B replication runs the identical battery and decoding configuration; its L1 rung is reconstructed from the memory arm's first round, whose message construction is identical to the L1 configuration (validated on Qwen3.5-27B, where the reconstructed baseline matches the directly-run L1 accuracies exactly on \texttt{reach} and \texttt{val} and within 6~pp on all tasks).

\paragraph{Ladder levels.}
\Cref{tab:app_oracle_ladder} summarizes the four levels; L1 and L2 reuse existing response matrices so that all levels share the same jointly calibrated $(\theta_1, \theta_2)$ space.

\begin{table}[t]
\centering
\setlength{\tabcolsep}{3pt}
\renewcommand{\arraystretch}{1.15}
\scriptsize
\caption{The oracle ladder. Stages of solving a structural item (encode, traverse, extract) are replaced by exact computation from right to left.}
\label{tab:app_oracle_ladder}
\begin{tabularx}{\columnwidth}{@{}*{4}{>{\raggedright\arraybackslash}X}l@{}}
\toprule
Level & Encode & Traverse & Extract & Data \\
\midrule
L1 Direct/CoT & LLM (in-head) & LLM (in-head) & LLM & existing \\
L2 Scratchpad & LLM & LLM + file tools & LLM & existing \\
L3 CTE & LLM (writes code) & machine execution & LLM & new \\
L4 Oracle injection & oracle (PDDL) & oracle enumerator & LLM & new \\
\bottomrule
\end{tabularx}
\end{table}

\paragraph{CTE arms.}
Beyond the paper's CoT and scratchpad baselines, L3 is instantiated with four arms: \texttt{cte\_uniform} (code compilation for all eight subtasks), \texttt{cte\_matched} (code for the Dimension~2 tasks \texttt{reach}, \texttt{land}, \texttt{areach}; CoT for the rest), \texttt{cte\_mismatched} (the reversed assignment, the critical ablation), and a \texttt{memory} arm that replays a Reflexion-style failure-analysis loop \citep{DBLP:conf/nips/ShinnCGNY23} as a generic-scaffold baseline.

\paragraph{Pre-registered hypotheses.}
\textbf{H1 (jump point):} structural accuracy is flat from L1 to L2 and jumps at L3 or L4, identifying the binding stage. \textbf{H2 (dose response):} accuracy and re-fitted $\theta_2$ increase monotonically from T0 to T3 while $\theta_1$ stays flat. \textbf{H3 (content causality):} distractor injections push accuracy below T0 at matched length and format. \textbf{H4 (dimension specificity):} gains concentrate on Dimension~2 items; per-item gain regresses positively on the continuous loading $b_2$ with task fixed effects; operational controls change minimally. \textbf{H5 (probe linkage):} items previously labeled latent absent in \Cref{sec:externalization_failures} are preferentially rescued by injection. \textbf{H6 (residual diagnosis):} failures remaining at T3 are predominantly extraction errors.

\paragraph{Enumeration artifacts.}
Full state spaces on these items are far too large to enumerate layer by layer (over $2\times10^{5}$ states at depth 8 on a logistics instance; much larger elsewhere), so evidence is computed at the level of reachable propositions and actions. Per item, a Fast Downward--based engine produces, for every candidate atom (for \texttt{reach} and \texttt{land}) or grounded action (for \texttt{areach}), either a replay-verified witness plan, a complete-search unsolvability proof, or an \emph{unresolved} status (never silently called unreachable); each entry carries its witness-plan length. For \texttt{land}, each candidate's landmark status is decided by an avoidance check (remove the atom's add effects and test goal reachability with a complete search, with a vacuity guard for unsolvable goals). All artifacts are validated against the benchmark's gold answers; this validation additionally showed that the gold ``no'' landmark set is not necessity-exact (e.g., a grid-domain atom we prove is a true landmark is listed under ``no''), so the gold gate requires only that gold ``yes'' landmarks be provable by our engine. One of 390 structural items carries a documented exception (a grid action whose status resisted a 600-second complete proof and is reported as unresolved rather than guessed).

\paragraph{Injection tiers.}
\Cref{tab:app_ladder_tiers} lists the four injection tiers. The T1 cutoff $k$ is calibrated per task as the median witness-plan length over reachable entries, so partial evidence covers roughly half of each item's reachable set. The verdict itself is never injected; tiers differ only in evidence content, and the T0/filler arms pad prompts with length-matched filler text so that ``more enumeration information'' is separated from ``more tokens.'' A format-sensitivity pilot (Qwen3.5-27B, T2, 60--80 items per structural task per format) compared a JSON serialization against a natural-language bullet-trace serialization; the bullet trace won on all three structural tasks (52.2\% vs.\ 43.3\% overall) and was frozen for the main runs.

\begin{table}[t]
\centering
\renewcommand{\arraystretch}{1.15}
\scriptsize
\caption{Injection tiers for the L4 oracle ladder. Evidence is injected, never the verdict.}
\label{tab:app_ladder_tiers}
\begin{tabularx}{\columnwidth}{@{}l>{\raggedright\arraybackslash}X@{}}
\toprule
Tier & Injected content \\
\midrule
T0 & None (equivalent to the L1 CoT prompt; length-matched filler variant as placebo) \\
T1 & Partial enumeration: reachable atoms/actions with witness-plan length $\le k$, $k$ = per-task median \\
T2 & Complete reachable proposition set (size-capped with summary fallback); reachable action set for \texttt{areach} \\
T3 & T2 + witness-plan lengths; for \texttt{land}, candidate landmark atoms with witness lengths and the goal atoms \\
\bottomrule
\end{tabularx}
\end{table}

\paragraph{Distractor control.}
Distractors are minimal perturbations of the true enumeration artifact that support the opposite verdict where possible (e.g., for a reachability-true item, a reachable set that excludes goal-satisfying states; for \texttt{land}, a candidate set missing a true landmark or including a spurious one). Every distractor is checked against the PDDL ground truth, the perturbation edit distance is logged, and implausible mutations are rejected; human spot-checks cover roughly 30 distractors.

\paragraph{CTE pipeline and error taxonomy.}
The \texttt{cte\_pipeline} consists of a code-generation prompt, an isolated executor (a locked-down container: no network, read-only filesystem, CPU/memory/PID/output caps, wall-clock timeout; the model's program runs as standard-library-only Python), a constrained answer format, and an automatic error classifier. Residual failures are classified as \texttt{compile\_failed} (no program produced), \texttt{timeout}, \texttt{resource\_error}, \texttt{extraction\_error} (the program's own printed verdict is correct but the final answer is wrong), and translation/enumeration error (everything else). A stratified sample of 100 programs is exported for human audit.

\paragraph{Analysis.}
New model--condition pairs enter the existing joint-calibration MIRT fit with shared item parameters; bootstrap confidence intervals are computed for $\Delta\theta_1$ and $\Delta\theta_2$ per level and tier. Item-level regressions of per-item gain $\Delta_j$ on $b_2$ include task fixed effects. L4 outcomes are cross-tabulated against the probe labels of \Cref{sec:externalization_failures} (latent absent vs.\ externalization failure).

\subsection{Additional Oracle-Ladder Results}
\label{sec:appendix_ladder_results}

This subsection reports the oracle-ladder details summarized in \Cref{sec:oracle_ladder,sec:manipulation}.

\paragraph{Dose response and controls (Qwen3.5-27B).}
On \texttt{reach}, accuracy moves from 55.4\% (T0) to 36.9\% (T1), 70.0\% (T2), and 70.0\% (T3). Partial evidence misleads, and complete evidence saturates. Re-fitted abilities follow the same shape with dimensional selectivity. The $\Delta\theta_2$ interval moves from $[-0.68, -0.08]$ (T1) to $[+0.22, +0.80]$ (T2) and $[+0.10, +0.67]$ (T3), while the $\Delta\theta_1$ interval spans zero at every tier. On the operational controls (\texttt{app}, \texttt{prog}, \texttt{val}), every true-versus-filler difference has a CI that spans zero.

\paragraph{Probe linkage.}
Cross-tabulating T2 outcomes against the probe labels of \Cref{sec:externalization_failures}, rescue rates for latent-absent and externalization-failure items are statistically indistinguishable (14.3\% vs.\ 12.2\%, Fisher $p = 1.0$). Injection does not preferentially rescue either representational class.

\paragraph{CTE error taxonomy.}
\Cref{tab:cte_taxonomy} breaks down the L3 outcomes. Self-compiled enumeration fails predominantly through translation and enumeration errors in the model's own programs, not through timeouts, resource limits, or verdict extraction.

\begin{table}[t]
\centering
\setlength{\tabcolsep}{3pt}
\scriptsize
\caption{CTE (L3) outcomes per task on Qwen3.5-27B: self-compiled enumeration fails predominantly through translation/enumeration errors in the model's own programs, not through timeouts, resource limits, or verdict extraction.}
\label{tab:cte_taxonomy}
\begin{tabular}{@{}lrrrr@{}}
\toprule
Task & \shortstack[r]{Compile\\fail} & Timeout & Resource & \shortstack[r]{Translation/\\enum.\ errors\\(of 130)} \\
\midrule
\texttt{reach}  & 4 & 3  & 11 & 68 \\
\texttt{land}   & 3 & 5  & 10 & 92 \\
\texttt{areach} & 0 & 8  & 5  & 107 \\
\texttt{prog}   & 5 & 1  & 0  & 107 \\
\bottomrule
\end{tabular}
\end{table}

\paragraph{Inference-time dissociation intervals.}
On Qwen3.5-27B, the \texttt{memory} arm raises both dimensions moderately ($\Delta\theta_1$ CI $[+0.29, +1.51]$, $\Delta\theta_2$ CI $[+0.45, +0.71]$), a profile unlike either structure-matched arm. On Gemma-4-12B, injection at T2 yields $\Delta\theta_2$ CI $[+0.47, +0.96]$ with the $\Delta\theta_1$ interval spanning zero at every tier. \texttt{cte\_matched} lowers $\theta_2$ ($[-1.11, -0.69]$), \texttt{cte\_uniform} collapses operational ability ($\Delta\theta_1$ CI $[-10.33, -0.16]$), and \texttt{memory} raises both dimensions ($\Delta\theta_1$ CI $[+0.13, +3.85]$, $\Delta\theta_2$ CI $[+0.38, +0.61]$).

\paragraph{Gemma-4-12B replication.}
\Cref{tab:ladder_gemma} repeats the battery on Gemma-4-12B, an independent family at roughly half the scale. Its L1 rung is reconstructed from the memory arm's first round (\Cref{sec:appendix_ladder}). On \texttt{reach}, T2 adds 14.7~pp over T0 (14.6~pp on Qwen). The dose response keeps its V shape ($-15.3$~pp at T1) and saturates at T2--T3. Distractor evidence drives \texttt{reach} below T0 ($-28.5$~pp, 95\% CI $[-39.0, -17.9]$), and filler is null. Gemma's CTE failures add budget exhaustion to translation errors. 51\% of its compile calls consume the entire 32{,}768-token reasoning budget (2.9\% on Qwen), and 35.3\% of items never produce a runnable program. Machine-supplied enumeration still rescues the same items (66.2\% at T2), so the model can consume a correct enumeration that it cannot produce. The \texttt{memory} arm again lifts operational tasks while leaving \texttt{areach} at floor.

\begin{table}[t]
\centering
\setlength{\tabcolsep}{3pt}
\scriptsize
\caption{Cross-family replication of \Cref{tab:ladder_results} on Gemma-4-12B (same arms and item sets; operational controls shown with L4-T2 injection for the specificity check).}
\label{tab:ladder_gemma}
\begin{tabular}{lcccc}
\toprule
Task & L1 CoT & L3 CTE & L4 (T2) & Memory \\
\midrule
\texttt{reach}  & 46.9\% & 16.9\% & \textbf{66.2\%} & 50.8\% \\
\texttt{land}   & 57.7\% & 13.1\% & 59.2\% & \textbf{66.2\%} \\
\texttt{areach} & 7.7\%  & 2.3\%  & \textbf{8.5\%}  & 8.5\% \\
\midrule
\texttt{app}    & 35.4\% & 6.9\%  & 33.8\% & 46.2\% \\
\texttt{prog}   & 77.7\% & 3.1\%  & 76.2\% & 84.6\% \\
\texttt{val}    & 93.1\% & 28.5\% & 91.5\% & 94.6\% \\
\bottomrule
\end{tabular}
\end{table}

\subsection{Competency-Aware Routing}
\label{sec:appendix_car}

The router maps each incoming item to an inference strategy from the item's predicted dimensional profile. Router inputs: (i) the oracle task label (scientific upper bound) and (ii) a lightweight classifier (TF-IDF + logistic regression) trained on item text with PDDL-domain-grouped cross-validation, so reported accuracy is out-of-fold on held-out domains and no gold labels are needed at deployment. Strategy sets: CoT for Dimension~1-predicted items with either CTE (the pre-registered assignment) or L4-T2 oracle injection (the deployable variant where a formal model is available) for Dimension~2-predicted items. Baselines: always-CoT, always-CTE, 1D difficulty routing (per-item mean baseline accuracy thresholds), and random routing. Costs are measured generation tokens per strategy (chars/4 estimate for reused baselines, logged where used). Success criterion: strict Pareto dominance over difficulty routing on the accuracy--cost frontier; we report both variants of the verdict.

\paragraph{Results.}
With CTE as the structural strategy, the oracle matched router reaches 54.8\% at 6.1k tokens, against 61.3\% at 2.4k for always-CoT and 33.0\% at 14.7k for always-CTE. With oracle injection as the structural strategy, the competency router reaches 63.3\%, against 61.3\% for always-CoT and 61.3\% for the surface router using CTE. The surface-feature classifier reproduces the full oracle gain (63.3\%). Competency routing strictly dominates 34 of 51 difficulty-routing points in the CTE variant and 1 of 51 in the injection variant. On Gemma-4-12B, the injection router reaches 63.7\% against 61.1\% for always-CoT at lower mean cost, the surface classifier keeps the full gain, competency routing strictly dominates 1 of 51 difficulty-routing points, and routing the hardest items to injection reaches 64.9\%.

\section{Supplementary Results from the Rebuttal Period}
\label{sec:appendix_rebuttal_results}

\subsection{Original Calibration and Model-Selection Context}
\label{sec:appendix_rebuttal_calibration}

The response to the sample-size concern reported 27 respondents (9 model checkpoints $\times$ 3 conditions) answering 1{,}040 items (8 tasks $\times$ 130 items). This is the respondent count of that original analysis, not a count for the expanded checkpoint list in \Cref{sec:models_and_conditions}. The task-aware decomposition, $b_{jd}=\alpha_{t(j),d}+\epsilon_{jd}$, separates eight interpretable task baselines per dimension (16 in 2D) from regularized item residuals. It does not remove the item residuals from the fitted parameterization or make the 1{,}040 task-inheriting labels independent.

The rebuttal used MIRT for comparative model selection rather than precise population-level recovery of each item's difficulty. The model-selection results in \Cref{tab:irt_aic} belong to that original calibration; they are not a new comparison fitted to every later diagnostic collection. The rebuttal described the first two PCA components as explaining 84\% of variance. We retain 84\% only as the textual claim made in the rebuttal: the original scree figure instead shows 61.9\% and 24.8\% (86.7\%, rounded to 87\%), while the updated summary in \Cref{fig:pca_scree} reports 88\%. The original textual and figure summaries were therefore inconsistent. PCA is an exploratory dimensionality check: orthogonal components do not, by themselves, establish a noncompensatory response mechanism. Likewise, the lower held-out NLL of the 4D model is not evidence of overfitting merely because it has more parameters; the 2D choice reflects the reported information criteria and interpretability tradeoff.

\subsection{Original Selective Neuron Analysis}
\label{sec:appendix_rebuttal_sna}

At Qwen~3.5~27B's cached layer~64, the rebuttal evaluated 5{,}120 neurons using Pearson correlations with the two difficulty coordinates and FDR correction at $\alpha=0.05$. It reported 2{,}629 $b_1$-selective and 2{,}343 $b_2$-selective neurons, with disjoint top-50 sets. The original top-50 prediction errors are retained in \Cref{tab:rebuttal_sna_mse}.

\begin{table}[htbp]
\centering
\footnotesize
\setlength{\tabcolsep}{3pt}
\caption{Rebuttal-period SNA MSE at layer~64. The updated fold-local analysis is in \Cref{tab:sna_mse}.}
\label{tab:rebuttal_sna_mse}
\begin{tabularx}{\columnwidth}{@{}>{\raggedright\arraybackslash}Xrr@{}}
\toprule
Target & Top-50 MSE & Full-layer MSE \\
\midrule
$b_1$ & 0.79 & 1.98 \\
$b_2$ & 2.53 & 3.65 \\
\bottomrule
\end{tabularx}
\end{table}

The earlier report found lower error for the top-50 subsets, but did not document repeating neuron selection within each training fold. The updated analysis in \Cref{sec:appendix_sna} explicitly performs fold-local selection and finds the opposite comparison: the full layer predicts both targets better. Consequently, the earlier sparse-subset advantage is not a current conclusion. Both analyses are exploratory and correlational. The dimension-selective assignment rule enforces disjoint sets, so disjointness is not independent validation and does not prove causally independent planning competencies or the multiplicative MIRT mechanism.

\subsection{Original In-Situ Statistical Robustness Suite}
\label{sec:appendix_rebuttal_isrs}

The rebuttal's In-Situ Statistical Robustness Suite (ISRS) compared the observed 2D noncompensatory NLL against 1{,}000 task-label permutations and 1{,}000 correctness-label shuffles. The observed NLL was below every resample in both tests, with reported one-sided $p=0.000999$ (\Cref{tab:rebuttal_null_tests}). These are the original NLLs, distinct from the updated values in \Cref{tab:null_tests}.

\begin{table}[htbp]
\centering
\footnotesize
\setlength{\tabcolsep}{3pt}
\caption{Rebuttal-period ISRS null tests. Both tests used 1{,}000 resamples and reported $p=0.000999$.}
\label{tab:rebuttal_null_tests}
\begin{tabularx}{\columnwidth}{@{}>{\raggedright\arraybackslash}Xrr@{}}
\toprule
Null & Mean NLL & Observed NLL \\
\midrule
Task-label permutation & 0.2065 & 0.1613 \\
Correctness-label shuffle & 0.4716 & 0.1613 \\
\bottomrule
\end{tabularx}
\end{table}

\begin{table}[htbp]
\centering
\footnotesize
\setlength{\tabcolsep}{3pt}
\caption{Rebuttal-period six-task collapsed-refit comparison. Accuracy is a proportion.}
\label{tab:rebuttal_task_collapsed}
\begin{tabularx}{\columnwidth}{@{}>{\raggedright\arraybackslash}Xrrr@{}}
\toprule
Refit & NLL & Acc. & $\Delta$ NLL \\
\midrule
Task-specific (6 tasks) & 0.1613 & 0.9611 & -- \\
Task-collapsed & 0.2284 & 0.9389 & +0.0671 \\
\bottomrule
\end{tabularx}
\end{table}

Benjamini--Hochberg correction at $\alpha=0.05$ retained 30 of 34 semantic-tag associations. \Cref{tab:rebuttal_fdr} records the six correlations and adjusted $q$-values printed in the rebuttal; their updated counterparts are in \Cref{tab:fdr_top}. These are item-level associations with task-inherited tags.

\begin{table}[htbp]
\centering
\footnotesize
\setlength{\tabcolsep}{3pt}
\renewcommand{\arraystretch}{1.12}
\caption{Rebuttal-period semantic-tag correlations after BH correction. Tag names follow \Cref{sec:appendix_semantic_codebook}.}
\label{tab:rebuttal_fdr}
\begin{tabularx}{\columnwidth}{@{}>{\raggedright\arraybackslash}Xcrr@{}}
\toprule
Tag & Dim. & $r$ & BH $q$ \\
\midrule
\path{requires_counterfactual_consistency} & $b_2$ & $-$0.645 & $4.8\!\times\!10^{-122}$ \\
\path{requires_search} & $b_2$ & +0.586 & $8.8\!\times\!10^{-96}$ \\
\path{verification_output} & $b_2$ & $-$0.586 & $8.8\!\times\!10^{-96}$ \\
\path{requires_goal_structure_tracking} & $b_2$ & +0.579 & $2.1\!\times\!10^{-93}$ \\
\path{requires_composition_over_steps} & $b_2$ & $-$0.552 & $3.5\!\times\!10^{-83}$ \\
\path{requires_goal_structure_tracking} & $b_1$ & $-$0.531 & $3.1\!\times\!10^{-76}$ \\
\bottomrule
\end{tabularx}
\end{table}

\subsection{Stronger Versus Lightweight Agent Scaffold}
\label{sec:appendix_rebuttal_scaffold}

The earlier Qwen~3.5~27B paired-subset experiment compared Direct, CoT, the original lightweight Tool Scratchpad, and a stronger OpenCode agent. The rebuttal described the stronger scaffold as supporting context-compaction cycles, search tools, and persistent session memory. \Cref{tab:rebuttal_scaffold_overall} preserves the four-condition comparison omitted from the updated three-condition table. OpenCode improved accuracy over Tool Scratchpad by 0.075 (7.5 percentage points), but remained below CoT.

\begin{table}[htbp]
\centering
\footnotesize
\setlength{\tabcolsep}{3pt}
\caption{Overall accuracy. Deltas are against the original Tool Scratchpad; dashes follow the rebuttal's reporting.}
\label{tab:rebuttal_scaffold_overall}
\begin{tabularx}{\columnwidth}{@{}>{\raggedright\arraybackslash}Xrr@{}}
\toprule
Condition & Acc. & $\Delta$ vs.\ Tool \\
\midrule
Direct & 0.468 & -- \\
CoT & 0.630 & +0.142 \\
Tool Scratchpad & 0.488 & -- \\
OpenCode Agent & 0.563 & +0.075 \\
\bottomrule
\end{tabularx}
\end{table}

Keeping the structural-difficulty quartile item sets fixed, OpenCode improved over the lightweight scaffold in every quartile (\Cref{tab:rebuttal_scaffold_b2}). Nevertheless, accuracy fell from 95.2\% in Q1 to 7.5\% in Q4. This is evidence of a persistent bottleneck \emph{under the tested scaffold}, not a universal impossibility of structural reasoning with tools. The updated analysis in \Cref{sec:appendix_scaffold} instead compares OpenCode with CoT on 800 paired items and permits only workspace file read/edit tools, rather than the compaction/search/memory configuration described in the rebuttal. The comparisons therefore differ in both configuration and collection and must not be pooled or read as the same run.

\begin{table}[htbp]
\centering
\footnotesize
\setlength{\tabcolsep}{3pt}
\caption{Rebuttal-period accuracy by fixed $b_2$ quartile. Tool denotes the original lightweight scratchpad.}
\label{tab:rebuttal_scaffold_b2}
\begin{tabularx}{\columnwidth}{@{}>{\raggedright\arraybackslash}Xrrr@{}}
\toprule
$b_2$ quartile & OpenCode & Tool & $\Delta$ \\
\midrule
Q1 (low) & 0.952 & 0.928 & +0.024 \\
Q2 & 0.829 & 0.759 & +0.071 \\
Q3 & 0.395 & 0.257 & +0.138 \\
Q4 (high) & 0.075 & 0.000 & +0.075 \\
\bottomrule
\end{tabularx}
\end{table}

\begin{samepage}
\paragraph{Scope of the additional evidence.}
The logistics failure example reported in the rebuttal is already retained in \Cref{sec:appendix_qual_example}. The analyses above add traceable controls and representation-level evidence within ACPBench-Hard/PDDL-style symbolic planning. They do not establish benchmark-independent planning cognition, rule out all floor/variance effects, supply unreported full-family probes and task-clustered inference, or report fold-level uncertainty for the dimensional-model comparison. The benchmark's designed subtask partition may contribute to the recovered axes. These limits are distinct from the targeted interventions and routing experiments reported in \Cref{sec:interventions,sec:appendix_interventions}.
\end{samepage}

\section{AI Assistance}
\label{sec:appendix_ai_assistance}

\paragraph{Manuscript preparation.}
An LLM was used only for polishing language; the scientific content is entirely original.

\paragraph{Implementation assistance.}
No coding agent was used in the previous ARR submission. In the current ARR resubmission, a coding agent was used to assist implementation. The experimental design was developed entirely by the authors.

\end{document}